\documentclass[10pt,twocolumn,letterpaper]{article}

\makeatletter
\@namedef{ver@everyshi.sty}{}
\makeatother

\usepackage{iccv}
\usepackage{times}
\usepackage{epsfig}
\usepackage{graphicx}
\usepackage{amsmath}
\usepackage{amssymb}
\usepackage{mathtools}
\usepackage{amsxtra}
\usepackage{booktabs}
\usepackage[noend]{algpseudocode}
\usepackage{multirow}
\usepackage{colortbl}
\usepackage{color}
\usepackage{xcolor}
\usepackage{wrapfig}
\usepackage{changes}
\usepackage{amsmath}
\usepackage{enumerate}
\usepackage{xcolor,soul,framed} %,caption
\usepackage{float}
\usepackage{hhline}
\usepackage{eso-pic}
\usepackage{soul}
\usepackage{enumerate}
\usepackage{enumitem}
\usepackage{paralist}
\usepackage[ruled]{algorithm2e}

\definecolor{Ocean}{RGB}{236,240,236}
\usepackage[pagebackref=true,breaklinks=true,colorlinks,bookmarks=false]{hyperref}

\iccvfinalcopy % *** Uncomment this line for the final submission

\def\iccvPaperID{****} % *** Enter the ICCV Paper ID here

\ificcvfinal\fi

\begin{document}

%%%%%%%%% TITLE
\title{UniDA3D: Unified Domain Adaptive 3D Semantic Segmentation Pipeline}

\author{Ben Fei$^{*,1}$, Siyuan Huang$^3$, Jiakang Yuan$^1$, Botian Shi$^2$,\\ Bo Zhang$^{\dagger,2}$, Weidong Yang$^{\dagger,1}$, Min Dou$^2$, Yikang Li$^2$\\
$^1$Fudan University $^2$Shanghai AI Laboratory, $^3$Shanghai Jiaotong University\\
{\tt\small {bfei21@m.fudan.edu.cn, siyuan\_sjtu@sjtu.edu.cn, jkyuan22@m.fudan.edu.cn}}
% \and
% Siyuan Huang\\
% Shanghai Jiaotong University\\
% \and
% Jiakang Yuan\\
% Fudan University\\
% \and
% Botian Shi, Bo Zhang, \\
% Shanghai AI Laboratory\\
}

\maketitle

\newcommand\blfootnote[1]{%
\begingroup
\renewcommand\thefootnote{}\footnote{#1}%
\addtocounter{footnote}{-1}%
\endgroup
}
% \blfootnote{{$\dagger$}Corresponding author. Email: wdyang@fudan.edu.cn,\\ bo.zhangzx@gmail.com}
\blfootnote{{$^*$}This work was done when  Ben Fei was an intern at Shanghai AI Laboratory.} 
\blfootnote{{$^\dagger$}Corresponding to: Bo Zhang (zhangbo@pjlab.org.cn), Weidong Yang (wdyang@fudan.edu.cn)}

\maketitle
\ificcvfinal\thispagestyle{empty}\fi

%%%%%%%%% ABSTRACT
\begin{abstract}
   State-of-the-art 3D semantic segmentation models are trained on off-the-shelf public benchmarks, but they will inevitably face the challenge of recognition accuracy drop when these well-trained models are deployed to a new domain. In this paper, we introduce a Unified Domain Adaptive 3D semantic segmentation pipeline (UniDA3D) to enhance the weak generalization ability, and bridge the point distribution gap between domains. Different from previous studies that only focus on a single adaptation task, UniDA3D can tackle several adaptation tasks in 3D segmentation field, by designing a unified source-and-target active sampling strategy, which selects a maximally-informative subset from both source and target domains for effective model  adaptation. Besides, benefiting from the rise of multi-modal 2D-3D datasets, UniDA3D investigates the possibility of achieving a multi-modal sampling strategy, by developing a cross-modality feature interaction module that can extract a representative pair of image and point features to achieve a bi-directional image-point feature interaction for safe model adaptation. Experimentally, UniDA3D is verified to be effective in many adaptation tasks including: 1) unsupervised domain adaptation, 2) unsupervised few-shot domain adaptation; 3) active domain adaptation. Their results demonstrate that, by easily coupling UniDA3D with off-the-shelf 3D segmentation baselines, domain generalization ability of these baselines can be enhanced. 
\end{abstract}

%%%%%%%%% BODY TEXT
\section{Introduction}
\label{sec:intro}

\begin{figure}[t]
% \vspace{-0.2cm}
  \centering
  \includegraphics[width=1.0\linewidth]{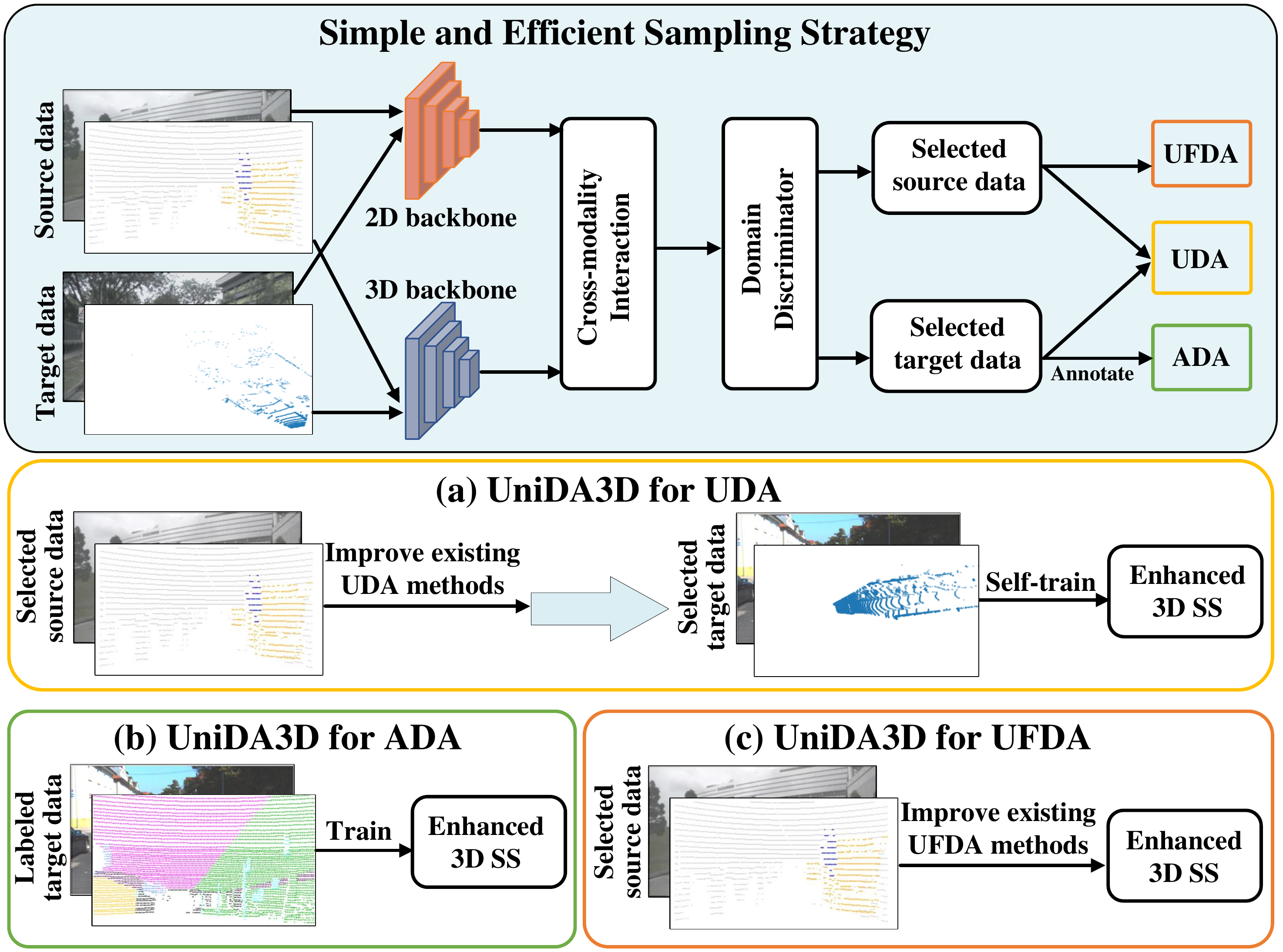}
  \caption{Our UniDA3D leverages a \textbf{unified} multi-modal sampling strategy that can effectively address various DA tasks: (a) UDA, (b) ADA, and (c) UFDA.}
\vspace{-0.3cm}
\end{figure} 

%3D semantic segmentation methods \cite{r7,r24,r33} often encounter the problem of shift or gap between different but related domains. The task of cross-modal domain adaptation (DA) \cite{r1} for 3D segmentation is designed to address the problem, inspired by 3D datasets usually containing 2D and 3D modalities. Similar to most DA tasks, labels here are only available in the source domain, whereas the target domain has no segmentation labels. 

In recent years, 3D semantic segmentation models\cite{graham20183d, wang2019graph, hu2021bidirectional} have achieved remarkable gains, owing to the large-scale annotated datasets, such as SemanticKITTI~\cite{behley2019semantickitti} and nuScenes~\cite{caesar2020nuscenes}, \textit{etc}. However, 3D semantic segmentation models still suffer from a severe performance drop issue when they are directly deployed to a novel domain. Actually, one straightforward method to alleviate such a performance drop issue is to build a specific dataset for the novel domain, by extensively collecting target-domain data and performing labor-intensive human annotation~\cite{sindagi2019ha, ding2021modeling}. But this is impractical in many real-world applications such as autonomous driving. 

Domain Adaptation (DA), as one of the typical techniques in the transfer learning community, aims to tackle the above-mentioned performance drop issue by learning domain-invariant representations~\cite{wilson2020survey, toldo2020unsupervised,nguyen2021domain}, which includes Unsupervised Domain Adaptation (UDA), Unsupervised Few-shot Domain Adaptation (UFDA), and Active Domain Adaptation (ADA) tasks, according to the condition of the target domain. Recently, inspired by UDA in 2D image~\cite{zhang2019category, bolte2019unsupervised, wang2020differential}, some researchers~\cite{jaritz2020xmuda,peng2021sparse, liu2021adversarial} try to address the 3D point cloud-induced domain variations under \textbf{a single adaptation task} such as the UDA setting. For example, xMUDA~\cite{jaritz2020xmuda} attempts to extract 2D and 3D features using two different network branches, exploiting the inter-modal feature complementarity. However, different from these UDA 3D segmentation works that focus on leveraging multi-modal data only under \textbf{a single adaptation task} (\textit{i.e.} UDA), we aim to develop a unified domain adaptive segmentation method for achieving a safe-and-unified 3D segmentation model transfer under \textbf{several adaptation tasks}.

To design a unified method that can be effective under different DA task settings, we resort to exploring a model-agnostic or task-agnostic strategy. We observe that, during the 3D model source-to-target adaptation process, there are many irrelevant source-domain samples and redundant target-domain samples. On the one hand, due to the intra-domain feature variations, some samples from the source domain may present a large data distribution difference from the target domain data. Learning from such irrelevant source-domain samples will be harmful to the model transfer towards the target domain, since the source-only model will fit the feature distribution that is quite different from the target domain, causing an \underline{irrelevant source-domain adaptation}. On the other hand, in autonomous driving scenarios, samples within the same target-domain sequence have a similar data distribution, which results in \underline{redundant target-domain adaptation} by self-training UDA methods~\cite{zou2018unsupervised, mei2020instance, zhang2021prototypical}. Thus, we propose a Unified Domain Adaptive 3D semantic segmentation (UniDA3D) pipeline, which can pick up a maximally-informative subset from both source and target domains to reduce the difficulty of model source-to-target adaptation.

Besides, our UniDA3D also can fully leverage the multi-modal data from different domains to perform a multi-modal balanced sampling process. During the model adaptation process, UniDA3D uses a cross-modality feature interaction module, which encodes features from a single image or point cloud modality, and then performs an image-to-point and point-to-image feature-level information interaction by a symmetrical cross-branch attention structure. Furthermore, the learned cross-modal features (Fig.~\ref{fig3}) are utilized as an important proxy for the subsequent sampling process, which is versatile to many DA tasks. As illustrated in Fig.~\ref{sec:intro}, such a subset sampling way can be easily combined with the off-the-shelf DA variants, achieving a better source-domain pre-trained performance and a promising target-domain adaptation performance.

The main contributions can be summarized as follows:
\begin{compactitem}
    \item[1.]
    We explore the possibilities of leveraging bi-domain cross-modal sampling method to achieve a unified domain adaptive 3D semantic segmentation.
    \item[2.]
    UniDA3D pipeline is proposed, which utilizes a cross-modality feature interaction module to fuse the feature’s multi-modal information, boosting the model's adaptation performance.
    \item[3.]
    UniDA3D can be deployed to many adaptation tasks simultaneously, which include: 1) UDA task where all data from the target domain are unlabeled; 2) UFDA task where we can only access few-shot unlabeled samples from the target domain; 3) ADA task where a portion of unlabeled target data is selected to be annotated by an oracle. Experiments are conducted on several public benchmarks, including nuScenes~\cite{caesar2020nuscenes}, A2D2~\cite{geyer2020a2d2}, SemanticKITTI~\cite{behley2019semantickitti}, and VirtualKITTI~\cite{jaritz2022cross}, and their results show that UniDA3D can be easily applied to multiple adaptation tasks, to enhance the model transferability, outperforming the existing works, xMUDA~\cite{jaritz2020xmuda}, AUDA~\cite{liu2021adversarial}, and DsCML~\cite{peng2021sparse} by $7.75\%$, $7.92\%$, and $4.61\%$.
\end{compactitem}

% \vspace{-0.10cm}
\section{Related Works}
% \vspace{-0.10cm}
\subsection{Active Domain Adaptation}

Active learning aims to develop annotation-efficient algorithms via sampling the most representative samples to be labeled by an oracle \cite{ren2021survey}. Most recently, active learning coupled with domain adaptation, termed as Active Domain Adaptation (ADA), has great practical significance. Nevertheless, only a few previous researchers focus on addressing the problem, pioneered by active adaptation in the area of sentiment classification for text data \cite{rai2010domain}. Rita et al. \cite{chattopadhyay2013joint} choose target samples to learn importance weights for source instances by solving a convex optimization problem of minimizing Maximum Mean Discrepancy (MMD). Recently, Su et al. \cite{su2020active} try to study ADA problem in the context of Convolution Neural Networks (CNN), and instances are selected based on their designed uncertainty and “targetness”. However, these sampling strategies are designed based on the 2D image domain, and it is intractable to directly apply these 2D image-based sampling strategies to the 3D image-point multi-modal task. In our work, for the first time, we design a novel active-and-adaptive segmentation baseline to sample the most informative 2D-3D pairs to enhance the weak cross-domain generalization ability of a well-trained 3D segmentation model.

\begin{figure*}[t]
  \centering
  \includegraphics[width=\linewidth]{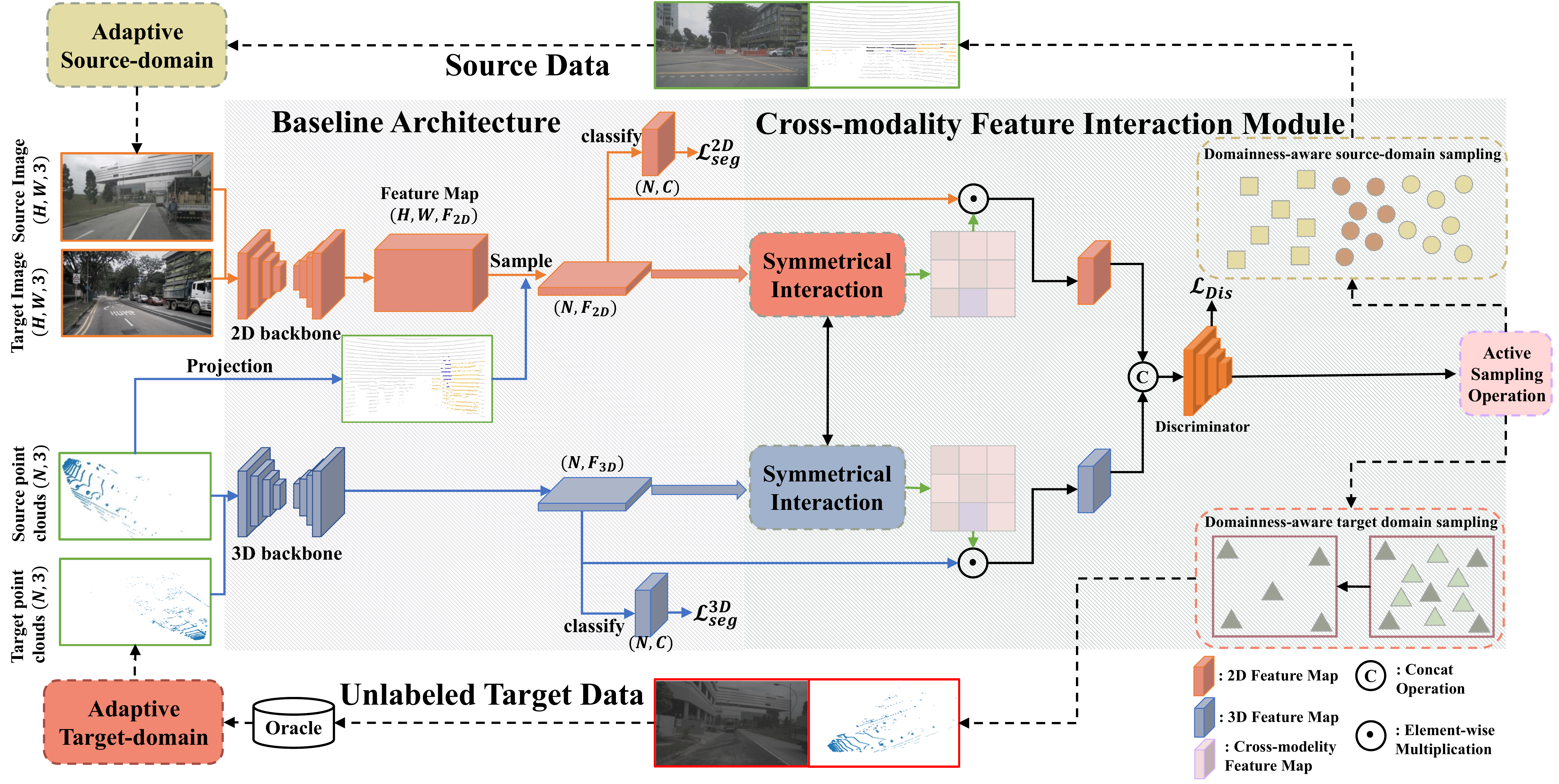}
  \caption{The network architecture of UniDA3D, which consists of a 3D segmentation baseline, the cross-modality feature interaction module, and the active sampling operation. The 3D segmentation baseline comprises a 2D U-Net-Style ConvNet \cite{he2016deep} backbone, which takes an image as input, and a 3D U-Net-Style SparseConvNet \cite{graham20183d} backbone, which receives a point cloud as input. Cross-modality feature interaction module can leverage the features ($F_{2D}$ and $F_{3D}$) to exploit a representative pair of image features and point cloud features to achieve a bi-directional image-point feature interaction. The last active sampling operation utilizes the interactive features to perform both the source-domain sampling and target-domain sampling via the unified module.}
\vspace{-0.3cm}
\label{figure2}
\end{figure*}

% \vspace{-0.10cm}
\subsection{General and UDA 3D Semantic Segmentation}
In 3D semantic segmentation, 3D point clouds are often represented as voxels. For instance, SSCNs~\cite{graham20183d} and the following works~\cite{choy20194d,hu2021vmnet,ye2021learning} leverage hash tables to convolve only on sampled voxels, allowing for very high resolution with typically only one point per voxel. Point-based methods conduct computation in continuous 3D space and can directly take point clouds as input. PointNet++~\cite{qi2017pointnet++} utilizes point-wise convolution and max-pooling to extract global features and local neighborhood aggregation for hierarchical learning with CNN. Following that, continuous convolutions~\cite{wang2018deep} and deformable kernels~\cite{thomas2019kpconv} have been proposed. DGCNN~\cite{wang2019dynamic} and LDGCNN~\cite{zhang2019linked} perform convolution on the edges of a point cloud. More recently, xMUDA~\cite{jaritz2020xmuda}, AUDA~\cite{liu2021adversarial}, and DsCML~\cite{peng2021sparse} utilize point clouds and images for UDA task in 3D semantic segmentation. However, these methods only focus on UDA task and overlook the intrinsic characteristics of the source and target domain. 
Therefore, in this work, we try to solve multiple adaptation tasks in a unified pipeline rather than a single adaptation task in 3D semantic segmentation~\cite{jaritz2020xmuda, liu2021adversarial, peng2021sparse}.
% In this work, SparseConvNet~\cite{graham20183d} is chosen as a 3D backbone, which is widely utilized in the UDA methods~\cite{jaritz2020xmuda, liu2021adversarial, peng2021sparse}.

\section{The Proposed Method}

Our method is proposed to enhance the adaptability of a 3D semantic segmentation model by assuming the presence of 2D images and 3D point clouds. In this section, we will first define the problem and illustrate the network architecture in Sec.~\ref{subsec_pre}. Following that, the UniDA3D pipeline is described in Sec.~\ref{sec:3.2}.

\subsection{Preliminary}
\label{subsec_pre}

\noindent\textbf{Problem Definition.} Suppose that $\mathcal{S}$ and $\mathcal{T}$ define a source domain and a target domain, both of which contain different modalities including 2D images and 3D point clouds $\{(\boldsymbol{x}^s_{2 \mathrm{D}}, \boldsymbol{x}^s_{3 \mathrm{D}})\}^{n_s}_{i=1}$ and $\{(\boldsymbol{x}^t_{2 \mathrm{D}}, \boldsymbol{x}^t_{3 \mathrm{D}})\}^{n_t}_{i=1}$. 
The purpose of cross-domain 3D semantic segmentation is to adapt a well-trained segmentation baseline from a labeled source domain $\mathcal{S}$ to a new target domain $\mathcal{T}$ with the data distribution shift.

To design a unified cross-domain 3D semantic segmentation pipeline under different target-domain conditions, we consider the following adaptation tasks: 1) Unsupervised Domain Adaptation (UDA), where we can access all unlabeled 2D-3D sample pairs from the target domain $\mathcal{T}$; 2) Unsupervised Few-shot Domain Adaptation (UFDA), where only a few data (\textit{e.g.,} $5\%$ target data) from the unlabeled target domain $\mathcal{T}$ are available; 3) Active Domain Adaptation (ADA), where one can sample a subset from the full set of the unlabeled target domain and perform the manual annotation process. An annotation budget $\mathcal{B}_{t}$ denotes the total amount of sampled target domain data for ADA task, while a budget $\mathcal{B}_{s}$ is denoted as the total number of sampled source domain data.

%with the corresponding 3D segmentation labels $\boldsymbol{y}^s_{3 \mathrm{D}}$

%We consider the following cross-domain 3D segmentation scenarios., containing the image $\boldsymbol{x}^t_{2 \mathrm{D}}$ and point cloud $\boldsymbol{x}^t_{3 \mathrm{D}}$. Furthermore, the images $\boldsymbol{x}^{2 \mathrm{D}}$ are of spatial size $(H, W, 3)$ while the point clouds $\boldsymbol{x}^{3 \mathrm{D}}$ of spatial size $(N, 3)$, with $N$ points in the camera field of view.}

\noindent\textbf{3D Segmentation Architecture.} Given annotations from the source domain, a segmentation baseline is trained in a supervised manner with cross-entropy loss function for the 2D image $\boldsymbol{x}^s_{2D}$ and 3D point cloud $\boldsymbol{x}^s_{3D}$ as follows:
\vspace{-0.15cm}
\begin{equation}
\begin{aligned}
    & \!\!\mathcal{L}^{seg}_{2D}\left(x^s_{2D}\!, \phi(y^s_{3D})\right)\!=\!-\frac{1}{N}\! \sum_{n=1}^N \!\sum_{c=1}^C \phi{(y^s_{(n, c)})} \!\log P_{2D}^{(n, c)},\\
    & \!\!\mathcal{L}^{\text{seg}}_{3D}\left(x^s_{3D}\!, y^s_{3D}\right)\!=\!-\frac{1}{N}\!\! \sum_{n=1}^N \!\sum_{c=1}^C y^s_{(n, c)} \!\log P_{3D}^{(n, c)} \text {,}
\end{aligned}
\label{eq1}
\end{equation}
\vspace{-0.15cm}

\noindent where $y^s_{(n, c)}$ and $P^{(n, c)}$ stand for the ground-truth label and prediction of the point $n$ for the class $c$, respectively. $\phi$ is the projection of the point cloud to the front view. Herein, the overall objective of the source domain can be formulated as:
\vspace{-0.15cm}
\begin{equation}\small
    \!\! \min _{\theta_{2D}, \theta_{3D}} \frac{1}{N^s} \sum_{x_s \in S} \mathcal{L}_{2D}^{seg}\left(x^s_{2D}, \phi(y^s_{3D})\right)+\mathcal{L}_{3D}^{seg}\left(x^s_{3D}, y^s_{3D}\right),
\label{eq2}
\end{equation}
\vspace{-0.15cm}

\noindent where $\theta_{2D}$ and $\theta_{3 D}$ are the parameters of the $2 \mathrm{D}$ sub-network and the $3 \mathrm{D}$ sub-network, respectively.

\begin{figure}[t]
    \centering
    \includegraphics[width=0.8\linewidth]{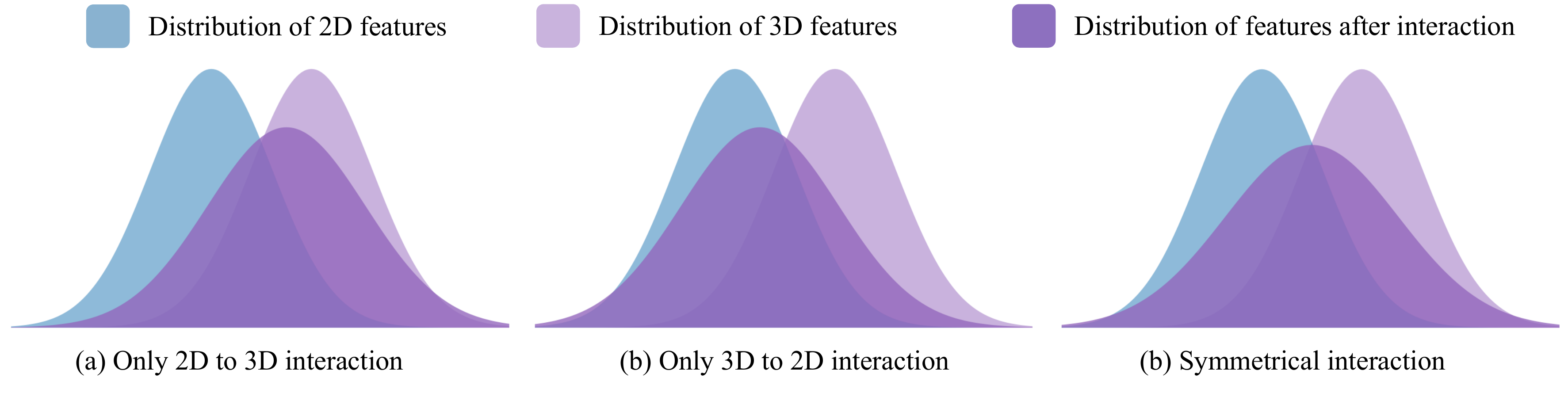}
    \caption{Illustration of unidirectional and symmetrical interaction. (a) The high-level features ($f_{2D}$, $f_{3D}$) undergo 2D to 3D interaction will obtain ($f_{2D}$, $\hat{f}_{3D}$), which samples data pairs largely by 3D features; (b) Otherwise, the high-level features ($f_{2D}$, $f_{3D}$) go through 3D to 2D interaction will get ($\hat{f}_{2D}$, $f_{3D}$) to sample data pairs taking 2D features into more consideration. (c) To this end, our symmetrical interaction provides a novel way to sample data pairs by considering both 2D and 3D features, leading to a more powerful sampling capability.}
\label{fig3}
\vspace{-0.2cm}
\end{figure}

\subsection{UniDA3D: Unified Domain Adaptive 3D Semantic Segmentation Pipeline}
\label{sec:3.2}

%Previous works mainly focus on cross-model learning within a single dataset, facing unforeseen cross-model performance hurt when they are directly used in cross-domain applications. 
Previous models~\cite{jaritz2020xmuda, liu2021adversarial, peng2021sparse} can only tackle a single adaptation task. As a pioneer work, UniDA3D tries to address multiple adaptation tasks in 3D segmentation field using a unified method and avoids the frequent adaptation method switching when facing different adaptation tasks or scenarios. UniDA3D is proposed to reduce the domain discrepancies from two aspects: 1) \underline{\textbf{Domain Active 3D Segmentation}}, meaning that we actively mine a subset of source-and-target data that are representative and transferable for dynamically changing target-domain distribution, and 2) \underline{\textbf{Domain Adaptive 3D Segmentation}}, meaning that we adapt a well-trained baseline to the target domain according to the above-sampled subset of the original data.

\noindent\textbf{\underline{Domain Active 3D Segmentation}: A Bi-domain Cross-modality Feature Interaction Module.} 
Although inter-domain differences exist, we found that many frames in the source domain have a similar data distribution to those in the target domain. This phenomenon motivates us to mine target-domain-like frames from the source domain to enhance the model adaptability, regarded as \textbf{source-domain sampling}. On the other hand, for 3D segmentation scenarios, there are many semantically-duplicated samples between adjacent frames. To this end, to reduce the cost of target data acquisition and annotation, we propose to select a maximally-informative subset of a given unlabeled target domain to perform the pseudo-label or manual annotation process regarded as \textbf{target-domain sampling}.

%there are intrinsic characteristics lying in both source and target domains.  These frames can be selected based on the active sampling strategy, and utilized to enhance the performance of domain adaptation. there are also some representative frames, which can be leveraged to further enhance the domain adaptation capability of the networks

In this part, we study how to design a unified sampling strategy to pick up samples from both domains, to reduce the feature gaps between domains and enhance the model transferability. An effective method to alleviate the inter-domain data distribution differences of 3D point clouds is leveraging the multi-modality residing in the 3D segmentation dataset. Motivated by this, a cross-modality feature interaction module is exploited to train a domain discriminator that can dynamically evaluate the representativeness of each sample from image-point cloud modalities. Given a pair of images and point clouds ($x_{2D}$, $x_{3D}$), the 2D and 3D backbones are respectively used to extract a pair of high-level features ($f_{2D}$, $f_{3D}$), where $f_{2D}\in \mathbb{R}^{N \times F_{2D}}$ and $f_{3D}\in \mathbb{R}^{N \times F_{3D}}$. Although the feature pairs ($f_{2D}$, $f_{3D}$) extracted by the backbone network contain rich semantic information of a single modal, the cross-modal semantic relations between the images and point clouds are not taken into consideration. 
%they lack semantic-level interaction between each other, and

Different from previous works \cite{jaritz2020xmuda,peng2021sparse, liu2021adversarial} that align cross-modal features using a well-designed $KL$ divergence loss, UniDA3D aims to exploit the relations between modalities by symmetrical attention \cite{zhang2022learning}. This approach achieves better cross-modality feature interaction and thus is beneficial to pick up data representative for both image-and-point modalities. The feature maps from each modality $f \in \mathbb{R}^{N \times {F_{2D}}}$ can be described as $f=\left[f^1, f^2, \ldots, f^{N}\right]$, where $f_{2D}^i \in \mathbb{R}^{F_{2D}}$ and $f_{3D}^i \in \mathbb{R}^{F_{3D}}$. In our implementation, the ${F_{2D}}$ \underline{is equal to} ${F_{3D}}$. 

% \vspace{-0.10cm}
Given cross-modal feature pairs $\left(f_{2D}, f_{3D}\right)$ and cross-modality feature interaction relations $\left(R_{3D \rightarrow 2D}, R_{2D \rightarrow 3D}\right)$ (detailed calculation can be found in the Appendix), the enhanced representations can be written as follows:
\vspace{-0.10cm}
\begin{equation}
\begin{split}
\hat{f}_{2D}=\operatorname{FFN}\left(\operatorname{Norm}\left(f_{2D} \odot R_{3D, 2D}\right)\right),\\
\hat{f}_{3D}=\operatorname{FFN}\left(\operatorname{Norm}\left(f_{3D} \odot R_{{2D}, {3D}}\right)\right),   
\end{split}
\end{equation}
\vspace{-0.10cm}

% \begin{equation}
%     \left\{\begin{array}{l}
%     \hat{f}_{2D}=\operatorname{MLP}\left(\operatorname{Norm}\left(f_{2D} \odot R_{3D, 2D}\right)\right) \\
%     \hat{f}_{3D}=M L P\left(\operatorname{Norm}\left(f_{3D} \odot R_{{2D}, {3D}}\right)\right)
% \end{array}\right.
% \end{equation}

\noindent where $\odot$ denotes an element-wise multiplication operation. And the FFN module allows the backbone to focus on the cross-branch modality discrepancies over the predicted semantically-similar regions. Overall, $\hat{f}_{2D}$ and $\hat{f}_{3D}$ are defined as the output features and will be concatenated and then fed into a well-designed domain discriminator, as illustrated in Fig. \ref{figure2}. 

Furthermore, a 3-layer fully-convolutional domain discriminator $\mathcal{D}$ with parameters $\theta_{\text{Dis}}$ is constructed, which takes the symmetrical-interaction features $[\hat{f}_{2D}, \hat{f}_{3D}]$ as input and is trained to distinguish the source data from the target ones. As shown in Fig.~\ref{fig3}, the symmetrical-interaction features can take both 2D and 3D features into consideration, benefiting the training of the domain discriminator. We label the source domain and the target domain as '0' and '1', respectively. Let $\mathcal{L}_\mathcal{D}$ represent the domain classification loss of the discriminator. The training objective of the discriminator can be written as follows:
\vspace{-0.10cm}
\begin{equation}
\begin{split}
    &\mathcal{L}_{Dis} = \min _{\theta_{Dis}} \frac{1}{N^s} \sum_{x^s_{2D},x^s_{3D}} \mathcal{L}_\mathcal{D}\left([\hat{f}_{2D}, \hat{f}_{3D}], 1\right)\\&+\frac{1}{N^t} \sum_{x^s_{2D},x^s_{3D}} \mathcal{L}_\mathcal{D}\left([\hat{f}_{2D}, \hat{f}_{3D}], 0\right).
\end{split}
\end{equation}
\vspace{-0.10cm}

Once the trained domain discriminator is obtained, we can leverage the output of the domain discriminator to score each frame, which represents the domainness of a sample belonging to the source or target domain (see Algorithm~\ref{algo}). Since the active sampling strategy is one general approach, we can use it for both the source and target domain sampling. For the source domain, the higher score means that the frame complies with the distribution of the target domain. For the target domain, when the frame possesses the more informative characteristic, it will get a higher score. After the scoring stage is finished, all frames are sorted by the calculated scores, and the budget $\mathcal{B}_s$ or $\mathcal{B}_t$ frames are chosen as the sampled source data and target data.

\begin{algorithm}[t]\small
\caption{Source and Target Sampling Strategy}
\label{alg:alg_1}
\hspace*{0.02in} {\bf Input:}
A pair of high-level features [$f_{2D}$, $f_{3D}$] extracted by 2D and 3D backbone. Domain discriminator $\mathcal{D}$ for scoring pairs of data, containing cross-modality feature interaction module $I_{cross}$, sampling budget $\mathcal{B}$.\\
\hspace*{0.02in} {\bf Output:} %算法的结果输出
The selected source or target set $Z$
\begin{algorithmic}[1]
\State [$\hat{f}_{2D}$, $\hat{f}_{3D}$] = $I_{cross}$[$f_{2D}$, $f_{3D}$]
\State $s_i = \mathcal{D}[\hat{f}_{2D}, \hat{f}_{3D}]$, where $s_i$ denotes as $i$-th frame data containing point cloud and image pair from the source or target domain
\State Sort $s_i$ in descending order
\State Sampling the data pairs according to the sorted $s_i$ until budget $\mathcal{B}_s$ or $\mathcal{B}_t$.
\State \Return Selected source or target set $Z$
\end{algorithmic}
\label{algo}
\end{algorithm}

\noindent\textbf{\underline{Domain Adaptive 3D Segmentation}: Effective Model Adaptation Strategy.} When an informative subset jointly sampled from both source and target domains is determined, an effective adaptation strategy is designed to fully leverage these samples to enhance 3D segmentation model's adaptability. In this part, we investigate two representative target-oriented adaptation strategies to explore how to perform an effective model transfer based on these informative samples.

\textit{1) Self-training using pseudo-labels:} Cross-modal learning is complementary to the pseudo-labeling technique, which is originally employed in semi-supervised and unsupervised domain adaptation tasks~\cite{li2019bidirectional,yen20223d}. In detail, given a pre-trained source model, the pseudo-labels can be obtained by selecting a portion of high-confidence predicted results according to the pre-trained model. It should be emphasized that the \textbf{Pseudo-Labeling methods (PL)} generate labels for the whole target domain data yet bring many label noises. By comparison, we can leverage the UniDA3D to pick up samples with high domainness scores. Then, we only pseudo-label those selected samples to perform the subsequent self-training process. Such a sampling-based pseudo-labeling method is termed as \textbf{Active Pseudo-Labeling (APL)}. 
%APL leverages only 5\% of target data, which greatly minimizes the cost of pseudo-labeling.
Based on those pseudo-labeled target frames, the model is further adapted to the target domain using the following segmentation loss:
\vspace{-0.15cm}
\begin{equation}\small
    \begin{aligned}
    \!\! & \min _\theta  {\left[\frac{1}{|\mathcal{S}|} \sum_{\boldsymbol{x}^s}\left(\mathcal{L}_{\mathrm{seg}}\left(\boldsymbol{x}^s, \boldsymbol{y}^s_{3 \mathrm{D}}\right)\right)\right.}  \left.+\frac{1}{|\mathcal{T}|} \sum_{\boldsymbol{x}^t}\left( \mathcal{L}_{\mathrm{seg}}\left(\boldsymbol{x}^t, \hat{\boldsymbol{y}}_{3 \mathrm{D}}\right)\right)\right],
    \end{aligned}
\end{equation}
\vspace{-0.15cm}

\noindent where $\hat{\boldsymbol{y}}_{3 \mathrm{D}}$ represents the pseudo-labels.

% and $\lambda_{\mathrm{PL}}$ is a \textcolor{red}{hyper-parameter} that achieves a trade-off between the pseudo-label segmentation loss and its original segmentation loss. 

% For clarity, we will refer to the xMUDA variant that uses additional self-training with pseudo-labels as xMUDA ${}_{\mathrm{PL}}$.

\textit{2) Off-the-shelf UDA techniques:} Furthermore, our sampling strategy can be easily integrated into the existing methods, such as xMUDA~\cite{jaritz2020xmuda}, AUDA\cite{liu2021adversarial}, and the state-of-the-art DsCML \cite{peng2021sparse}, to further boost their cross-domain segmentation accuracy. Specifically, we first utilize the proposed sampling strategy to pick up the maximally-informative samples from both source and target domains and construct a new subset. Then, these UDA-based techniques are employed to adapt the baseline model using the newly-constructed subset to achieve a more effective and safe model adaptation. 

% \subsection{Overall Loss Function}
% \label{sec:3.3}

% % \noindent\textbf{Loss Function.} 

% \begin{equation}
%     \mathcal{L} = \mathcal{L}_{seg}^{2D}\left(x_s, y_s\right)+ \mathcal{L}_{\text {seg}}^{3D}\left(x_s, y_s\right) + \mathcal{L}_{Dis}
% \end{equation}

% \noindent\textbf{Adaptation Strategy.}

\begin{table*}[htbp]
\vspace{-0.05cm}
\centering
\resizebox{0.98\textwidth}{!}{%
\begin{tabular}{c|c|cc>{\columncolor{Ocean}}c|cc>{\columncolor{Ocean}}c|cc>{\columncolor{Ocean}}c|cc>{\columncolor{Ocean}}c}
\toprule[1.0pt]
                                        &                           & \multicolumn{3}{c|}{ \textbf{USA/Singapore} } & \multicolumn{3}{c|}{ \textbf{Day/Night} }                                                          & \multicolumn{3}{c}{ \textbf{A2D2/SemanticKITTI} }          & \multicolumn{3}{c}{ \textbf{V-KITTI/SemanticKITTI} }                                        \\ \cline{3-14} 
\multirow{-2}{*}{}                      & \multirow{-2}{*}{ \textbf{Methods} } & 2D      & 3D      & Softmax Avg.   & 2D                          & 3D                          & Softmax Avg.                & 2D                          & 3D                          & Softmax Avg.  & 2D                          & 3D                          & Softmax Avg.               \\ \midrule[0.8pt]
                                        & Oracle                    & 66.4    & 63.8    & 71.6           & 48.6                        & 47.1                        & 55.2                        & 58.3                        & 71.0                        & 73.7 
                                     & 66.3                        & 78.4                        & 80.1 
                                     \\
                                        & MinEnt (CVPR-19)\cite{vu2019advent}                    & 53.4    & 47.0    & 59.7           & {\color[HTML]{000000} 44.9} & {\color[HTML]{000000} 43.5} & {\color[HTML]{000000} 51.3} & {\color[HTML]{000000} 38.8} & {\color[HTML]{000000} 38.0} & {\color[HTML]{000000} 42.7} 
                                        & {\color[HTML]{000000} 37.8} & {\color[HTML]{000000} 39.6} & {\color[HTML]{000000} 42.6}
                                        \\
                                        %& Deep logCORAL (ICLR-18) \cite{r12}             & 52.6    & 47.1    & 59.1           & {\color[HTML]{000000} 41.4} & {\color[HTML]{000000} 42.8} & {\color[HTML]{000000} 51.8} & {\color[HTML]{000000} 35.8} & {\color[HTML]{000000} 39.3} & {\color[HTML]{000000} 40.3} \\
                                        & PL (CVPR-19) \cite{li2019bidirectional}                        & 55.5    & 51.8    & 61.5           & 43.7                        & 45.1                        & 48.6                        & 37.4                        & 44.8                        & 47.7
                                         & 21.5                        & 44.3                        & 35.6 
                                        \\
                                        & CyCADA (ICML-18) \cite{hoffman2018cycada}                    & 54.9    & 48.7    & 61.4           & 45.7                        & 45.2                        & 49.7                        & 38.2                        & 44.3                        & 43.9  
                                        & -                        & -                        & -
                                        \\
                                        & AdaptSegNet (CVPR-18) \cite{tsai2018learning}              & 56.3    & 47.7    & 61.8           & 45.3                        & 44.6                        & 49.6                        & 38.8                        & 44.3                        & 44.2
                                        & -                        & -                        & -
                                        \\
                                        & CLAN (CVPR-19) \cite{luo2019taking}                      & 57.8    & 51.2    & 62.5           & 45.6                        & 43.7                        & 49.2                        & 39.2                        & 44.7                        & 44.5
                                        
                                        & -                        & -                        & - 
                                        \\
                                        & xMUDA (CVPR-20) \cite{jaritz2020xmuda}                     & 59.3    & 52.0    & 62.7           & 46.2                        & 44.2                        & 50.0                        & 36.8                        & 43.3                        & 42.9

                                        & 42.1                        & 46.7                        & 48.2
                                        \\
                                        & xMUDA + PL (CVPR-20) \cite{jaritz2020xmuda}                     & 61.1    & 54.1    & 63.2           & 47.1                        & 46.7                        & 50.8                        & 43.7                        & 48.5                        & 49.1 
                                        & 45.8                        & 51.4                        & 52.0 
                                        \\
                                        & AUDA \cite{liu2021adversarial}                      & 59.7    & 51.7    & 63.0           & 48.7                        & 46.2                        & 55.7                        & 43.3                        & 43.3                        & 47.3 & 41.4                        & 47.7                        & 49.3 \\
                                        & AUDA + PL \cite{liu2021adversarial}                      & 59.8    & 52.0    & 63.1           & 49.0                        & 47.6                        & 54.2                        & 43.0                        & 43.6                        & 46.8       & 41.6        & 53.1       & 51.2                  \\
                                        & DsCML (CVPR-21) \cite{peng2021sparse}                     & 61.3    & 53.3    & 63.6           & 48.0                        & 45.7                        & 51.0                        & 39.6                        & 45.1                        & 44.5    &  46.2       &  42.9      &   49.9                     \\
& DsCML + CMAL (CVPR-21) \cite{peng2021sparse}              & 63.4    & 55.6    & 64.8           & 49.5                        & 48.2                        & 52.7                        & 46.3                        & 50.7                        & 51.0  &    47.6     & 43.1       &  51.2
\\
\multirow{-12}{*}{\begin{tabular}[c]{@{}c@{}}Domain adaptation \\ methods\end{tabular}} & DsCML + CMAL + PL (CVPR-21) \cite{peng2021sparse}              & 63.9    & 56.3    & 65.1           & 50.1                        & 48.7                        & 53.0                        & 46.8                        & 51.8                        & 52.4 &    48.1                    & 44.4                       & 52.6 \\ \midrule[0.8pt]
                                        & Source   only             & 53.4   & 46.5   & 61.3          & 42.2                       & 41.2                       & 47.8                       & 34.2                       & 35.9                       & 40.4 
                                        
                                         & 26.8                       & 42.0                       & 42.2 
                                        \\
      & Source only + Source-domain sampling                      & 55.2   & 49.7   & 63.1          & 45.4                       & 40.8                       & 51.7                       & 35.1                       & 38.0                       & 44.7  & 33.0         & 44.2        & 45.1 \\   & \textit{\textbf{Improvement}}  & \textit{\textbf{+1.8}}   & \textit{\textbf{+3.2}}   & \textit{\textbf{+1.8}}          & \textit{\textbf{+3.2}}                      & \textit{\textbf{+0.4}}                       & \textit{\textbf{+3.9}}                       & \textit{\textbf{+0.9}}                       & \textit{\textbf{+2.1}}                       & \textit{\textbf{+4.3}} & \textit{\textbf{+6.2}}                       & \textit{\textbf{+2.2}}                       & \textit{\textbf{+2.9}}  \\
      & Source only + Source-domain sampling + PL                      & 59.4   & 53.6   & 63.3          & 46.2                       & 42.6                       &   51.4                      & 42.0                      & 35.9                       & 45.8 & 34.4                      & 44.7                      & 47.5 \\   & \textit{\textbf{Improvement}}  & \textit{\textbf{+5.0}}   & \textit{\textbf{+7.1}}   & \textit{\textbf{+2.0}}          & \textit{\textbf{+4.0}}                      & \textit{\textbf{+1.4}}                       & \textit{\textbf{+3.6}}                       & \textit{\textbf{+7.8}}                       & \textit{\textbf{+5.4}}                       & \textit{\textbf{+5.4}} & \textit{\textbf{+7.6}}                       & \textit{\textbf{+2.7}}                       & \textit{\textbf{+5.3}} \\
      & Source only + Source-domain sampling + APL                      & 60.9   & 53.1   & 66.8          & 47.0                       & 42.2                       & 51.9                       & 41.9                       & 41.4                       & 46.7 &36.2                       & 40.8                       & 46.8 \\ \multirow{-7}{*}{\begin{tabular}[c]{@{}c@{}}Sampling source-domain \\ w/ UniDA3D\end{tabular}}  & \textit{\textbf{Improvement}}  & \textit{\textbf{+7.5}}   & \textit{\textbf{+6.6}}   & \textit{\textbf{+5.5}}          & \textit{\textbf{+4.8}}                      & \textit{\textbf{+1.0}}                       & \textit{\textbf{+3.1}}                       & \textit{\textbf{+7.7}}                       & \textit{\textbf{+5.5}}                       & \textit{\textbf{+6.3}}    & \textit{\textbf{+10.2}}                       & \textit{\textbf{-1.2}}                       & \textit{\textbf{+4.6}}                \\ \midrule[0.8pt]
                                        & xMUDA + Source-domain sampling                & 55.9   & 50.1   & 63.4          & 45.6                       & 42.2                       & 51.2                       & 43.6                       & 47.6                       & 50.8    &  44.0       &  50.8      &   54.8                   \\ & \textit{\textbf{Improvement}}                & \textit{\textbf{-}}   & \textit{\textbf{-}}  & \textit{\textbf{+0.7}}          & \textit{\textbf{-}}                       & \textit{\textbf{-}}                       & \textit{\textbf{+1.2}}                       & \textit{\textbf{-}}                      & \textit{\textbf{-}}                       & \textit{\textbf{+7.9}}    & \textit{\textbf{-}}                      & \textit{\textbf{-}}                       & \textit{\textbf{+2.6}}                    \\ & xMUDA + Source-domain sampling + PL               & 60.8   & 51.4   & 63.6          & 46.2                       & 42.6                       & 51.4                       & 42.7                       & 43.3                       &  50.6   &   42.5      &  51.4      &    56.6                   \\ & \textit{\textbf{Improvement}}                & \textit{\textbf{-}}   & \textit{\textbf{-}}  & \textit{\textbf{+0.6}}          & \textit{\textbf{-}}                       & \textit{\textbf{-}}                       & \textit{\textbf{+1.4}}                       & \textit{\textbf{-}}                      & \textit{\textbf{-}}                       & \textit{\textbf{+1.5}}    & \textit{\textbf{-}}                      & \textit{\textbf{-}}                       & \textit{\textbf{+4.6}}                     \\ & xMUDA + Source-domain sampling + APL                & 57.6  & 54.6   & 63.8          & 48.4                       & 42.8                       & 51.5                       & 45.1                       & 46.3                       & 50.6   &  47.4       &   50.8     &  56.7                     \\ & \textit{\textbf{Improvement}}                & \textit{\textbf{-}}   & \textit{\textbf{-}}  & \textit{\textbf{+0.6}}          & \textit{\textbf{-}}                       & \textit{\textbf{-}}                       & \textit{\textbf{+0.7}}                       & \textit{\textbf{-}}                      & \textit{\textbf{-}}                       & \textit{\textbf{+1.5}}  & \textit{\textbf{-}}                      & \textit{\textbf{-}}                       & \textit{\textbf{+4.7}}                      \\
                                        & AUDA + Source-domain sampling                 & 54.7   & 50.3   & 63.2          & 49.1                       & 48.4                       & 54.4                       & 43.7                       & 42.3                       & 47.0     &  46.2       &  52.7      &  54.3                   \\ & \textit{\textbf{Improvement}}                & \textit{\textbf{-}}   & \textit{\textbf{-}}  & \textit{\textbf{+0.0}}          & \textit{\textbf{-}}                       & \textit{\textbf{-}}                       & \textit{\textbf{+0.2}}                       & \textit{\textbf{-}}                      & \textit{\textbf{-}}                       & \textit{\textbf{+0.2}}  & \textit{\textbf{-}}                      & \textit{\textbf{-}}                       & \textit{\textbf{+5.0}}                       \\ & AUDA + Source-domain sampling + PL                 & 59.4   & 53.59   & 63.3          & 49.1                       & 41.3                       & 54.0                       & 42.6                       & 37.9                       & 47.1    &  48.3       &   52.5     &   54.2                   \\ & \textit{\textbf{Improvement}}                & \textit{\textbf{-}}   & \textit{\textbf{-}}  & \textit{\textbf{+0.1}}          & \textit{\textbf{-}}                       & \textit{\textbf{-}}                       & \textit{\textbf{-0.2}}                       & \textit{\textbf{-}}                      & \textit{\textbf{-}}                       & \textit{\textbf{+0.3}}  & \textit{\textbf{-}}                      & \textit{\textbf{-}}                       & \textit{\textbf{+3.0}}                      \\
                                        & AUDA + Source-domain sampling + APL                 & 58.3   & 52.4   & 63.6          & 48.5                       & 41.6                       & 54.7                       & 42.0                       & 39.7                       & 47.9    &  45.7       & 53.7       &   54.9                   \\ & \textit{\textbf{Improvement}}                & \textit{\textbf{-}}   & \textit{\textbf{-}}  & \textit{\textbf{+0.5}}          & \textit{\textbf{-}}                       & \textit{\textbf{-}}                       & \textit{\textbf{-0.5}}                       & \textit{\textbf{-}}                      & \textit{\textbf{-}}                       & \textit{\textbf{+1.1}}  
                                        & \textit{\textbf{-}}                      & \textit{\textbf{-}}                       & \textit{\textbf{+3.7}} \\
            & DsCML +CMAL + Source-domain sampling                & 55.6   & 52.0   & 65.0          & 49.3                       & 41.6                       & 53.5                       & 43.5                       & 46.3                       & 51.1   &   52.9      & 49.5       & 53.0  \\ & \textit{\textbf{Improvement}}                & \textit{\textbf{-}}   & \textit{\textbf{-}}  & \textit{\textbf{+0.2}}          & \textit{\textbf{-}}                       & \textit{\textbf{-}}                       & \textit{\textbf{+0.8}}                       & \textit{\textbf{-}}                      & \textit{\textbf{-}}                       & \textit{\textbf{+0.1}} & \textit{\textbf{-}}                      & \textit{\textbf{-}}                       & \textit{\textbf{+1.8}} \\& DsCML +CMAL + Source-domain sampling + PL                & 59.9   & 55.2   & 65.7          & 49.1                       & 41.5                       & 53.3                       & 44.7                       & 47.0                       & 52.7  &   52.6       &  51.1      & 56.5 \\ & \textit{\textbf{Improvement}}                & \textit{\textbf{-}}   & \textit{\textbf{-}}  & \textit{\textbf{+0.6}}          & \textit{\textbf{-}}                       & \textit{\textbf{-}}                       & \textit{\textbf{+0.3}}                       & \textit{\textbf{-}}                      & \textit{\textbf{-}}                       & \textit{\textbf{+0.3}} & \textit{\textbf{-}}                      & \textit{\textbf{-}}                       & \textit{\textbf{+5.3}}  \\ & DsCML +CMAL + Source-domain sampling + APL                & 61.8   & 55.7   & 66.2          & 49.2                       & 42.6                       & 53.4                       & 47.3                       & 46.1    &  54.3       &  52.3      &  53.0                      &  56.7\\ \multirow{-16}{*}{\begin{tabular}[c]{@{}c@{}}UDA task w/ UniDA3D\end{tabular}} & \textit{\textbf{Improvement}}                & \textit{\textbf{-}}   & \textit{\textbf{-}}  & \textit{\textbf{+1.1}}          & \textit{\textbf{-}}                       & \textit{\textbf{-}}                       & \textit{\textbf{+0.4}}                       & \textit{\textbf{-}}                      & \textit{\textbf{-}}                       & \textit{\textbf{+1.9}}  & \textit{\textbf{-}}                      & \textit{\textbf{-}}                       & \textit{\textbf{+5.5}}                                             \\ \midrule[0.8pt]
                                        & UFDA ($10\%$ target samples)          & 55.0       & 49.3       & 63.0              & 45.1                           & 41.0                           & 51.3                           & 35.0                           & 37.9                           & 44.2  & 31.7        &   43.8     &   44.3                        \\
\multirow{-2}{*}{ UFDA task w/ UniDA3D}           & UFDA ($5\%$ target samples)         & 54.6       & 48.9       & 62.2              & 43.7                           & 40.7                           & 50.5                           & 34.5                           & 37.2                           & 43.6   &    30.9     &  40.2      &   40.7      \\ \bottomrule[1.0pt]
\end{tabular}%
}
\vspace{0.10cm}
\caption{Segmentation results (mIoU) for different adaptation tasks including UDA task, UDA methods coupled with UniDA3D, and UFDA task, using four multi-modal domain adaptation scenarios. ``\;V-KITTI\;'' denotes the VirtualKITTI dataset, and ``\;Avg.\;'' represents the performance which is obtained by computing the mean of the predicted 2D and 3D probabilities after softmax operation. ``\;PL\;'' denotes the pseudo-labeling operation for all target data, while ``\;APL\;'' is proposed to pseudo-label the actively sampled target-domain data.}
\label{table1}
\vspace{-0.15cm}
\end{table*}

\section{Experiments}

\subsection{Dataset Description}

Our UniDA3D is evaluated under 3 real-to-real adaptation tasks and 1 virtual-to-real adaptation task. The first task is the day-to-night adaptation, which presents a domain gap caused by the illumination changes, where the laser beams are almost invariant to illumination conditions while the camera suffers from a low-light environment. The second task is the country-to-country adaptation, representing a large domain difference where the 3D shapes and images might change frequently. The third task is the dataset-to-dataset adaptation, containing the variations in the sensor setup. The last task is the virtual-to-real adaptation with huge domain gaps existing in it. The widely-used autonomous driving datasets nuScenes \cite{caesar2020nuscenes}, A2D2~\cite{geyer2020a2d2}, and SemanticKITTI \cite{behley2019semantickitti} are utilized, where the LiDAR and camera are synchronized and calibrated to obtain the projection between a 3D point and its corresponding 2D image pixel. All these datasets consist of 3D annotations. And only the front camera images and the LiDAR points are utilized for simplicity and consistency across datasets. For nuScenes, the point-wise labels for 3D semantic segmentation are obtained by assigning the corresponding object label if that point lies inside an annotated 3D bounding box. If not, that point will be labeled as background. Following the works~\cite{jaritz2020xmuda,peng2021sparse}, two scenarios of Day/Night and USA/Singapore are conducted. For the adaptation from A2D2 to SemanticKITTI, the point-wise labels on A2D2 are directly provided, and the 10 categories shared between the two datasets are selected. To compensate for source/target class mismatch (e.g., VirtualKITTI/SemanticKITTI) or accommodate for class scarcity, a custom class mapping is applied, which is detailed in Appendix. Note that VirtualKITTI provides depth maps so that we can simulate LiDAR scanning via the uniform point sampling.

\begin{figure*}[t]
\vspace{0.15cm}
  \centering
  \includegraphics[width=\linewidth]{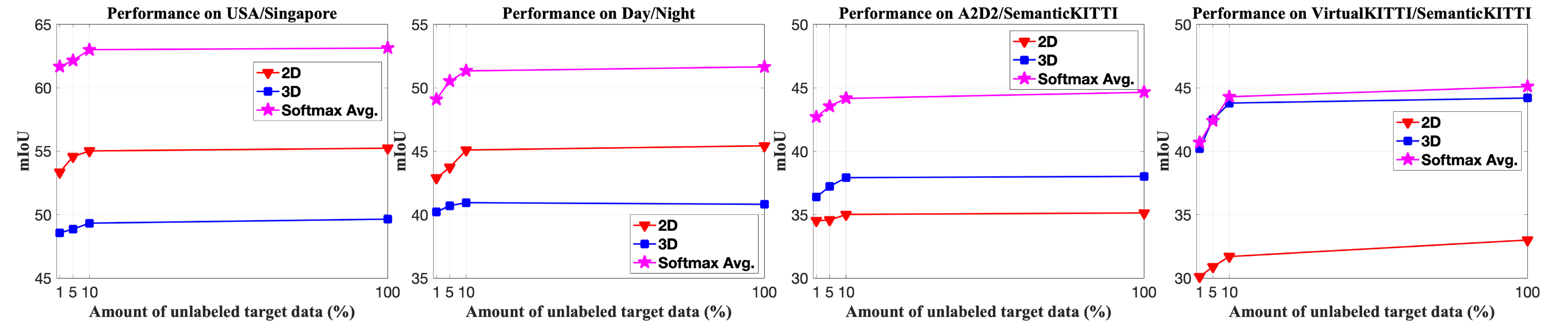}
  \caption{Segmentation results (mIoU) under the UFDA task, where we study the adaptation performance under different amounts of samples from the target domain: $100\%$, $10\%$, $5\%$, and $1\%$ target domain samples.}
\vspace{-0.1cm}
\label{fig:ufda}
\end{figure*} 

\subsection{The Designed Baselines}

\begin{table*}[t]
% \vspace{-0.25cm}
\small
\centering
\resizebox{\textwidth}{!}{%
\begin{tabular}{c|c|cc>{\columncolor{Ocean}}c|cc>{\columncolor{Ocean}}c|cc>{\columncolor{Ocean}}c|cc>{\columncolor{Ocean}}c}
\toprule[1.0pt]
\multirow{2}{*}{}    & \multirow{2}{*}{ \textbf{Methods} }                                              & \multicolumn{3}{c|}{ \textbf{USA/Singapore} } & \multicolumn{3}{c|}{ \textbf{Day/Night} } & \multicolumn{3}{c}{ \textbf{A2D2/SemanticKITTI} }& \multicolumn{3}{c}{ \textbf{VirtualKITTI/SemanticKITTI} } \\ \cline{3-14} 
                     &                                                                       & 2D      & 3D      & Softmax Avg.   & 2D     & 3D     & Softmax Avg. & 2D        & 3D       & Softmax Avg.  & 2D        & 3D       & Softmax Avg.    \\ \midrule[0.8pt]
                     & Source   only                                                         & 53.4   & 46.5   & 61.3          & 42.2  & 41.2  & 47.8        & 34.2     & 35.9    & 40.4  & 26.8                       & 42.0                       & 42.2          \\ \midrule[0.8pt]
\multirow{3}{*}{UDA} & xMUDA + PL \cite{jaritz2020xmuda}                                                                 & 61.1    & 54.1    & 63.2           & 47.1   & 46.7   & 50.8         & 43.7      & 48.5     & 49.1   & 45.8     & 51.4   & 52.0         \\
                     & AUDA + PL \cite{liu2021adversarial}                                                                 & 59.8    & 52.0    & 63.1           & 49.0   & 47.6   & 54.2         & 43.0      & 43.6     & 46.8    & 41.6     & 53.1  &  51.2       \\
                     & DsCML + CMAL +PL \cite{peng2021sparse}                                                                 & 63.9    & 56.3    & 65.1           & 50.1   & 48.7   & 53.0         & 46.8      & 51.8     & 52.4   &  47.6    &  43.1 & 51.2          \\ \midrule[0.8pt]
\multirow{4}{*}{ADA} & Ours + Source-domain sampling                                                                  & 55.2   & 49.7   & 63.1          & 45.4  & 40.8  & 51.7        & 35.1     & 38.0    & 44.7  & 33.0         & 44.2        & 45.1           \\
                     & Ours + Random sampling (Target domain) & 55.5   & 58.7   & 64.9          & 47.5  & 41.0  & 53.0        & 42.8         & 43.1        & 48.2    & 45.6        & 52.1        & 57.9            \\
                     & Ours + Source-domain sampling + Target-domain sampling       & 63.6   & 52.8   & 67.4          & 47.1  & 41.5  & 53.6        & 43.9         & 47.8        & 51.1  & 49.3         & 52.1        & 58.5   \\
                     & Ours + Source-domain sampling + Target-domain sampling + APL      & 62.7   & 56.3   & 68.1          & 49.8  & 41.2  & 54.6        & 46.2         & 45.7        & 52.8  & 51.8     &  53.7 &   59.2                  \\ \midrule[1pt]
                     & Oracle                                                                & 66.4    & 63.8    & 71.6           & 48.6   & 47.1   & 55.2         & 58.3      & 71.0     & 73.7    & 66.3                        & 78.4                        & 80.1         \\ \bottomrule[1.0pt]
\end{tabular}%
}
\vspace{0.05cm}
\caption{Segmentation results (mIoU) for the ADA task under $5\%$ target-domain annotation budget $\mathcal{B}_t$, where Random sampling denotes that we randomly sample $5\%$ target domain to perform the manual annotation.}
\label{table:ada}
\vspace{-0.2cm}
\end{table*}

In order to validate the effectiveness of our UniDA3D, 7 baselines are designed:
\textbf{Baseline 1} only leverages 2D features extracted by the 2D backbone. The 2D scores obtained by the domain discriminator $\mathcal{D}$ are used to sample the source and target data.
\textbf{Baseline 2} utilizes the 3D features for sampling, where the sampling process will depend on the 3D scores from the $\mathcal{D}$.
\textbf{Baseline 3} naively averages the 2D and 3D scores for the sampling strategy. Since the scores are calculated from a single modality, the sampled data might be distributed in a single modality. For instance, the data sampled by the 2D scores will concentrate more on the image information and vice versa.
\textbf{Baseline 4} integrates the 2D active domain adaptation method CLUE \cite{chen2005clue}, to be compared with our UniDA3D in 3D semantic segmentation.
\textbf{Baseline 5} further applies the LabOR \cite{shin2021labor} (a representative method for active segmentation model) to pick up data from the target domain.
\textbf{Baseline 6} uses 2D to 3D interaction to obtain the $\hat{f}_{3D}$, while \textbf{Baseline 7} utilizes 2D to 3D interaction to obtain the $\hat{f}_{2D}$. $\hat{f}_{2D}$ and $\hat{f}_{3D}$ will be leveraged for scoring by the domain discriminator $\mathcal{D}$.
The module-wise ablation study of all these baselines can be found in Table \ref{module-ablation}, and please refer to the Appendix for the network visualization of all designed baselines.
%\textbf{Ours.} To tackle the problems in the above-mentioned baselines, we propose the , which leverages a cross-attention module to enhance the relationship between the 2D and 3D features. The enhanced 2D and 3D features make both the images and point clouds into consideration. In this way, the scores from the trained discriminator can be utilized to sample the most representative data. 

\subsection{Implementation Details}
\noindent\textbf{Network Backbone.}
For a fair comparison with the state-of-the-art multi-modal 3D domain adaptation frameworks, we leverage the ResNet34~\cite{he2016deep} pre-trained on ImageNet \cite{ganin2015unsupervised} as the encoder for the 2D network branch and the SparseConvNet \cite{graham20183d} with U-Net architecture for 3D network branch. Moreover, the voxel size is set to 5 cm in the 3D network to ensure that only one 3D point lies in a single voxel. The models are trained and evaluated with PyTorch toolbox \cite{r8}. We train UniDA3D with one Tesla A100 GPU.

\noindent\textbf{Parameter Settings.} In the training stage, the batch size is set to 8, and the Adaptive Moment Estimation (Adam) \cite{r9} is used as the optimizer with $\beta_1$ = 0.9 and $\beta_2$ = 0.999. We set the learning rate of $1 \times e^{-3}$ initially and follow the poly learning rate policy \cite{chen2017rethinking} with a poly power of 0.9. We set the max training iteration as 100k. 

\subsection{Experimental Results}

\begin{table*}[t]
\vspace{-0.05cm}
\small
\centering
\resizebox{\textwidth}{!}{%
\begin{tabular}{c|cc>{\columncolor{Ocean}}c|cc>{\columncolor{Ocean}}c|cc>{\columncolor{Ocean}}c|cc>{\columncolor{Ocean}}c}
\toprule[1.0pt]
\multirow{2}{*}{ \textbf{Methods} } & \multicolumn{3}{c|}{ \textbf{USA/Singapore} } & \multicolumn{3}{c|}{ \textbf{Day/Night} } & \multicolumn{3}{c}{ \textbf{A2D2/SemanticKITTI} } & \multicolumn{3}{c}{ \textbf{VirtualKITTI/SemanticKITTI} } \\ \cline{2-13} 
                         & 2D      & 3D      & Softmax Avg.   & 2D     & 3D     & Softmax Avg. & 2D        & 3D       & Softmax Avg.   & 2D        & 3D       & Softmax Avg.   \\ \midrule[0.8pt]
Baseline1: UniDA3D-2D                  & 55.2   & 47.1   & 62.8          & 45.2  & 40.9  & 50.6        & 34.8     & 36.6    & 43.0    & 36.2    & 40.6   &  44.9      \\
Baseline2: UniDA3D-3D                  & 53.8   & 48.8   & 61.8          & 44.9  & 41.2  & 51.0        & 35.5     & 38.4    & 42.7     &  31.7   & 40.8 &   44.6    \\
Baseline3: UniDA3D-2D-3D-Avg               & 53.4   & 48.9   & 61.8          & 45.0  & 41.2  & 51.1        & 35.7     & 36.9    & 43.2     & 35.2     &  40.5 &  45.0    \\
Baseline4: UniDA3D-CLUE \cite{chen2005clue}              & 53.2   & 46.3   & 60.1          & 42.8  & 41.0  & 49.4        &  35.2    & 36.0   & 40.6    &   31.2   &  40.4  &   42.7     \\
Baseline5: UniDA3D-LabOR \cite{shin2021labor}               & 52.2   & 50.0   & 58.9          & 44.0  & 40.1  & 49.2        &  31.3    & 32.0   & 38.6 &   33.1   & 40.1  & 43.4 \\
Baseline6: UniDA3D-2D-to-3D-Interaction           & 57.9   & 49.7   &   60.7         & 43.0  & 41.8   &  50.3       &  34.4    &  36.1   &  42.1   &  32.4    & 44.6   &   42.8     \\
Baseline7: UniDA3D-3D-to-2D-Interaction            & 56.8   & 46.5   & 62.2         & 45.3  & 42.2  &  51.3       &   35.0   &  35.0   &   42.1  &  35.2   & 40.8  & 44.3     \\
\midrule[0.8pt]
Ours: UniDA3D-Symmetrical-Interaction     & 55.2   & 49.7   & 63.1          & 45.4  & 40.8  & 51.7        & 35.1     & 38.0    & 44.7  &    33.0     &  44.2      &  45.1        \\ \bottomrule[1.0pt]
\end{tabular}%
}
\vspace{0.05cm}
\caption{The module-level ablation studies of UniDA3D under four multi-modal domain adaptation scenarios.}
\label{module-ablation}
\vspace{-0.35cm}
\end{table*}

In this part, we conduct experiments on UDA, UFDA, and ADA tasks, and comprehensively show the generality and effectiveness in addressing 3D segmentation model's domain discrepancies.

\noindent\textbf{Results on 3D UDA task.} 
\noindent\textit{1) The effectiveness of source-domain sampling}: As shown in Table~\ref{table1}, when the segmentation baseline model is deployed to a new domain (\textit{e.g.}, from USA to Singapore), its segmentation accuracy is seriously degraded (only $61.3\%$ compared with $71.6\%$ achieved by fully-supervised model). One cost-free solution provided by our UniDA3D is to use the designed source-domain sampling strategy, which can actually sample some target-domain-like source data to bridge the large domain shift from different modal data, achieving 1.8$\%$, 3.9$\%$, 4.3$\%$, and 2.9$\%$ accuracy gains for different cross-domain scenarios. The accuracy gain achieved by only source sampling strategy even surpasses some representative uni-modal UDA methods such as MinEnt~\cite{vu2019advent}, PL~\cite{li2019bidirectional}, CyCADA~\cite{hoffman2018cycada}, AdaptSegNet~\cite{tsai2018learning}, and CLAN~\cite{luo2019taking}. 

Furthermore, we would like to emphasize that another advantage of our UniDA3D is that it can be plugged and played into the existing UDA models (\textit{e.g.}, xMUDA~\cite{jaritz2020xmuda}, AUDA~\cite{liu2021adversarial}, and DsCML~\cite{peng2021sparse}), to further strengthen these UDA models' adaptability. For example, xMUDA coupled with our UniDA3D achieves $0.7\%$, $1.2\%$, $7.9\%$, and $2.6\%$ segmentation accuracy gain compared with xMUDA itself, for the USA/Singapore, Day/Night, A2D2/SemanticKITTI, and VirtualKITTI/SemanticKITTI scenarios. Also, we conduct extensive validations by inserting our UniDA3D into many UDA models, such as AUDA~\cite{liu2021adversarial} and DsCML~\cite{peng2021sparse}, and observe consistent segmentation accuracy gains.

\noindent\textit{2) The effectiveness of target-domain sampling}: APL means that we only perform the pseudo-labeling process for the selected target samples by UniDA3D. Also, it can be seen from Table~\ref{table1} that, compared with the widely-used Pseudo-Labeling (PL) strategy, APL can obtain a relatively high mIoU in A2D2/SemanticKITTI setting. This is mainly because the selected target data (SemanticKITTI) by UniDA3D presents less noise and can be pseudo-labeled more accurately. By merging the merits of the APL and self-training strategy, UniDA3D can further enhance the adaptability of the model.

\noindent\textbf{Results on 3D UFDA task.} Motivated by the outstanding performance of UniDA3D in UDA task, we try to reduce the number of unlabeled target data to check the domain-related representation ability of the designed discriminator with the cross-modality feature interaction module. Such a task setting can be regarded as UFDA task. Specifically, we randomly select a small number of target data (\textit{e.g.} $10\%$, $5\%$, and even $1\%$) to train the discriminator, and the results are shown in Fig.~\ref{fig:ufda}. It can be seen that owing to the robust source-domain sampling strategy, only a few-shot unlabeled target data also can effectively train the designed discriminator and achieve a comparable target-domain segmentation accuracy, providing an option to eliminate the need for unlabeled target data.

\noindent\textbf{Results on 3D ADA task.} In the above DA tasks, all samples from the target domain are unlabeled. Although performing the APL or PL based on self-training methods on the unlabeled target domain can improve the target-domain segmentation accuracy, there is still a large accuracy gap between the unsupervised target-domain model and the fully-supervised one. To this end, in this work, we also study the ADA task, which assumes that a portion of unlabeled target data selected by an active learning algorithm can be annotated by an oracle or human expert. Instead of using pseudo-labeled target samples, we utilize an oracle to annotate all selected target samples to investigate if our UniDA3D can be applied to ADA task.

\begin{figure}[t]
\vspace{-0.25cm}
  \centering
  \includegraphics[width=0.7\linewidth]{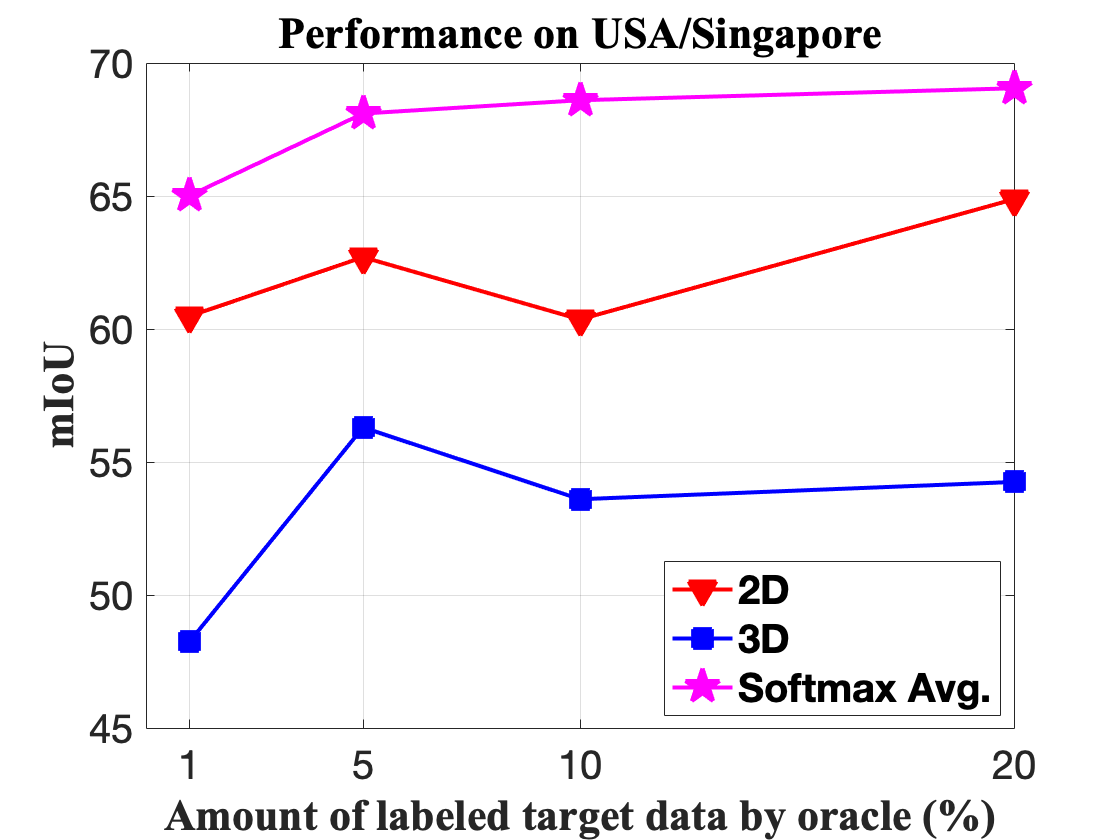}
  \vspace{-0.10cm}
  \caption{The influence of different target-domain annotation budget $\mathcal{B}_t$ ($20\%$, $10\%$, $5\%$, and $1\%$) on the ADA task. Here, we take the USA-to-Singapore setting for an example.}
\vspace{-0.15cm}
\label{budget}
\end{figure}

As shown in Table~\ref{table:ada}, the baseline model fine-tuned on the sampled source data and annotated target data can significantly enhance the model transferability, obtaining $67.4\%$, $53.6\%$, $51.1\%$, and $58.4\%$ segmentation accuracies under different settings. Furthermore, we conduct experiments on changing the budget of the target-domain annotation. The results can be found in Fig.~\ref{budget}, and we observe that the model performance in the target domain is improved with the increase of the annotation budget. We also observe that the $5\%$ annotation budget can be regarded as a good trade-off between the annotation cost and the model adaptability.

%If the baseline with APL is fine-tuned with selected target data, the results could enhance to 68.1, 54.6, and 52.8, respectively. It is noted that the adaptive baseline is superior to xMUDA and AUDA in the USA/Singapore and better than xMUDA and DsCML in the Day/Night. Moreover, our method performs better than the xMUDA, AUDA, and DsCML in A2D2/SemanticKITTI setting. 
%  enhances from 65.0 to 68.1 for the 1\% and 5\% budget, while the further increase the budget only brings a minor improvement in the results
%These results demonstrate that our active sampling strategy can choose the most representative target samples, significantly boosting the performance of the ADA task.

\subsection{Further Analyses}

\noindent\textbf{Module-level Ablation Studies.} To validate the effectiveness of the symmetrical attention module, ablation studies are conducted to gain insight into the symmetrical attention module. As shown in Table~\ref{module-ablation}, the sampled source data by only the 2D scores tend to boost the accuracy of baseline on 2D metric, while the samples selected only by 3D scores are more beneficial to improve the 3D segmentation results. When the mean sampling strategy (\textbf{Baseline 3}) is utilized, the average accuracy is able to be improved. By employing the designed symmetrical attention module, satisfactory overall results can be obtained, showing the effectiveness and strength of our symmetrical attention-based sampling strategy.

% \noindent\textbf{Active: how to effectively utilize the source-data.}
% It should be emphasized that, active source domain can boost the UFDA performance
% And analyse why an effective Active-and-Adaptive method can reduce the high-dependence of cross-domain segmentation model over the extensive unlabeled target samples. 

\noindent\textbf{Segmentation Accuracy for Both Domains.}
To validate the multi-domain generalization ability of our UniDA3D, further experiments are conducted to observe the segmentation accuracy of the model on both source and target domains. As shown in Table~\ref{ablation2}, the source-only baseline, which means that the model is trained only on the source domain and directly tested on the target domain, has the best segmentation accuracy for the source domain, but it is hard to be generalized to the target domain. On the other hand, the xMUDA baseline (UDA model) achieves better results on the target domain but not performs well on the source domain. Our UniDA3D performs a fine-tuning process on the selected source data (labeled) and the target data (pseudo-labeled or annotated), achieving better generalization on both the source and target domains.

\begin{table}[t]\small
\centering
\resizebox{0.98\linewidth}{!}
{
\begin{tabular}{cccc}
\toprule[1.0pt]
            & \begin{tabular}[c]{@{}c@{}} \textbf{Source} \\ 2D / 3D / Avg.\end{tabular} & \begin{tabular}[c]{@{}c@{}} \textbf{Target} \\ 2D / 3D / Avg.\end{tabular} & \begin{tabular}[c]{@{}c@{}} \textbf{Average} \\ 2D / 3D / Avg.\end{tabular} \\ \midrule[0.8pt]
Source only & \textbf{53.4 / 46.5 / 61.3}                                     & 31.4 / 43.4 / 49.1                                              & 42.4 / 45.0 / 55.2                                               \\
xMUDA       & 36.6 / 43.8 / 48.6                                              & 55.9 / 50.1 / 63.4                                              & 46.3 / 47.0 / 56.0                                               \\
UniDA3D        & 47.8 / 43.1 / 54.9                                              & 63.6 / 52.8 / 67.4                                              & \textbf{55.7 / 48.0 / 61.2}                                      \\ \bottomrule[1.0pt]
\end{tabular}
}
\vspace{0.10cm}
\caption{Generalization ability of UniDA3D. The results tested on both source and target domains are reported. Avg. denotes the softmax average performance.}
\vspace{-0.30cm}
\label{ablation2}
\end{table}

\section{Conclusion}

In this paper, we have presented a UniDA3D to tackle many adaptation tasks, including UDA, UFDA, and ADA tasks, by means of a unified pipeline. By the merits of the designed bi-domain cross-modality feature interaction module, UniDA3D can fully leverage features from different modalities to achieve an informative sample selection process from different domains. Experiments are conducted on widely-used 3D semantic segmentation datasets, showing the superiority of UniDA3D in boosting the segmentation model transferability.

\section*{Acknowledgement}
This work is supported by Science and Technology Commission of Shanghai Municipality (grant No. 22DZ1100102).

{\small
\bibliographystyle{ieee_fullname}
\bibliography{egbib}
}

\clearpage
\newpage

\appendix

\section{Details on Cross-modality Feature Interaction}

\setcounter{figure}{4}
\begin{figure}[htbp]
  \centering
  \includegraphics[width=0.97\linewidth]{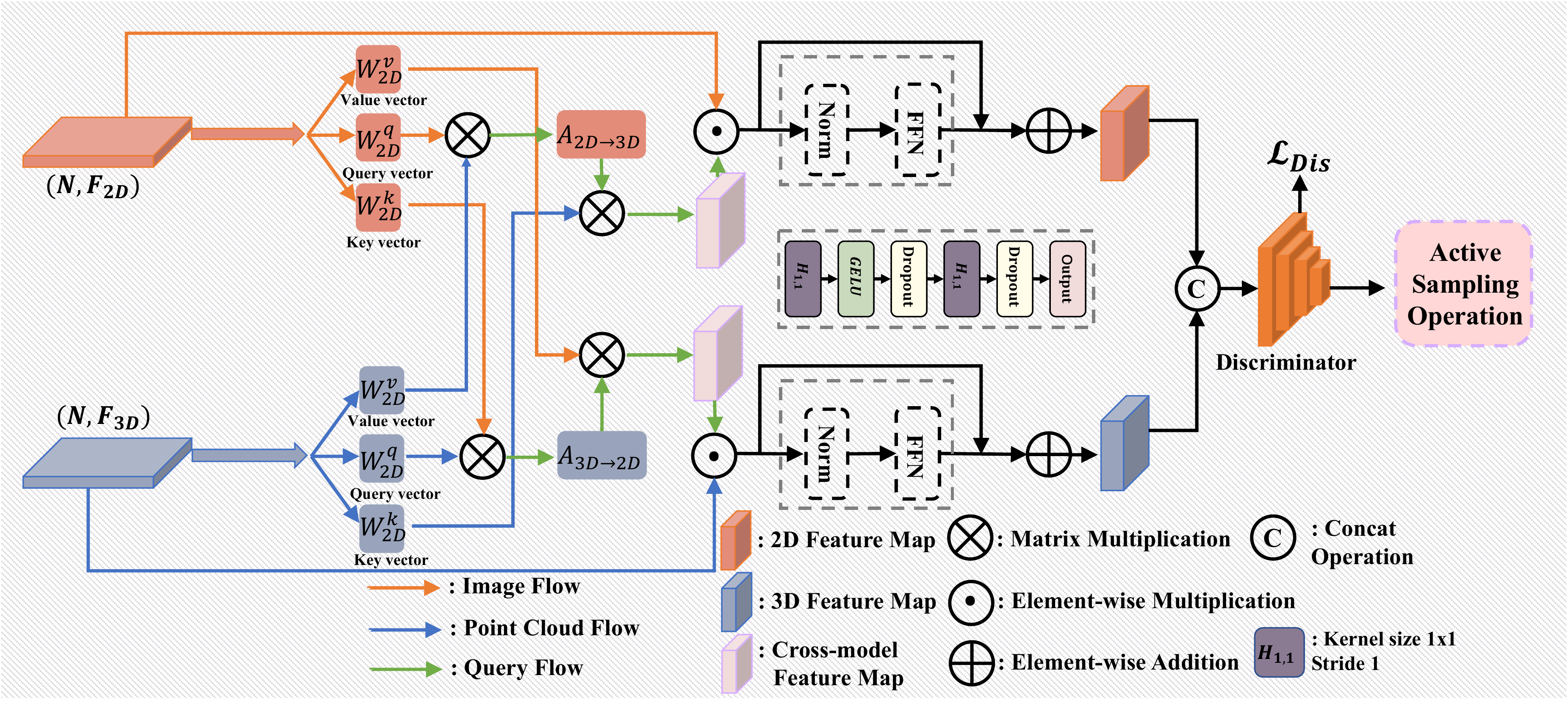}
  \vspace{0.05cm}
  \caption{The illustration of cross-modality feature interaction module.}
\label{ablation-module}
\vspace{0.15cm}
\end{figure}

Given feature maps with weight parameters $W_{2D}^q, W_{2D}^k, W_{2D}^v$ for the image branch $2D$, and parameters $W_{3D}^q, W_{3D}^k, W_{3D}^v$ for point cloud branch $3D$, the query vector $q^i$, key vector $k^i$, and value vector $v^i$ can be calculated as follows:
% \vspace{-0.15cm}
\begin{equation}\small
    \!\!{q_{2D}^{i} := W_{2D}^{q}f_{2D}^{i}} \text{,\quad}{k_{2D}^{i} := W_{2D}^{i}f_{2D}^{i}}\text{,\quad}{v_{2D}^{i} := W_{2D}^{v}f_{2D}^{i}},
\label{eq4.1}
\end{equation}
% \vspace{-0.25cm}
\begin{equation}\small
    \!\!q_{3D}^i := W_{3D}^q f_{3D}^i\text{,\quad}k_{3D}^i := W_{3D}^i f_{3D}^i\text{,\quad} 
    v_{3D}^i := W_{3D}^v f_{3D}^i.
\label{eq4.2}
\end{equation}

% \begin{equation}
%     \left\{\begin{array} { c } 
% {q_{2D}^{i} = W_{2D}^{q}f_{2D}^{i}} \\
% {k_{2D}^{i} = W_{2D}^{i}f_{2D}^{i}} \\
% {v_{2D}^{i} = W_{2D}^{v}f_{2D}^{i} }
% \end{array} \quad \left\{\begin{array}{c}
% q_{3D}^i=W_{3D}^q f_{3D}^i \\
% k_{3D}^i=W_{3D}^i f_{3D}^i \\
% v_{3D}^i=W_{3D}^v f_{3D}^i
% \end{array}\right.\right.
% \label{eq4}
% \end{equation}

After that, the symmetrical cross-attention is leveraged in a bi-direction manner: 1) The 2D branch-related features are obtained using the value vector $v_{3D}^i$ from $3D$ backbone branch, formulated as $\mathbf{3D} \rightarrow \mathbf{2D}$; 2) Similarly, the 3D branch-related features are acquired using the value vector $v_{2D}^i$ of the 2D backbone branch, formulated as $\mathbf{2D} \rightarrow \mathbf{3D}$.

Specifically, for $\mathbf{3D} \rightarrow \mathbf{2D}$, $\mathbf{A}_{3D \rightarrow 2D} \in \mathbb{R}^{N \times N}$ denotes the attention score matrix obtained via the matrix multiplication as follows:

% \vspace{-0.10cm}
\begin{equation}
    \mathbf{A}_{3D \rightarrow 2D}=K_{2D} V_{3D}^{\mathrm{T}},
\label{eq5}
\end{equation}

\noindent where $K_{2D}=\left[k_{2D}^1, \ldots, k_{2D}^{N}\right] \in \mathbb{R}^{N \times F_{2D}}$ and $V_{3D}=\left[v_{3D}^1, \ldots, v_{3D}^{N}\right] \in$ $\mathbb{R}^{N \times F_{3D}}$, which can be obtained through Eq.~\ref{eq4.1}, Eq.~\ref{eq4.2}. Moreover, a softmax layer combined with a scaling operation is leveraged to carry out the normalization for attention scores and find the semantically-related regions according to information from another modality $3D$ branch, which can be calculated as follows:

\begin{equation}
    R_{3D \rightarrow 2D}=\operatorname{Softmax}\left(\frac{\mathbf{A}_{3D \rightarrow 2D}}{\sqrt{F_{2D}}}\right) V_{2D},
\label{eq6}
\end{equation}

\noindent where $V_{2D}=\left[v_{2D}^1, v_{2D}^2, \ldots, v_{2D}^{N}\right] \in \mathbb{R}^{N \times F_{2D}}$, and $R_{3D \rightarrow 2D} \in \mathbb{R}^{N \times F_{2D}}$ represents the encoded semantic relations from 3D to 2D features. After that, the $R_{3D \rightarrow 2D}$ is reshaped to the same dimension as the backbone features $f_{2D} \in \mathbb{R}^{F_{2D} \times N}$ and then fused with $f_{2D}$ by an element-wise addition, in order to enhance the semantically similar backbone features.

For $\mathbf{2D} \rightarrow \mathbf{3D}$, the $R_{2D \rightarrow 3D} \in \mathbb{R}^{N \times F_{3D}}$ can be easily obtained by performing a symmetrical process described above, which can be written as follows:
% \vspace{-0.10cm}
\begin{equation}
\begin{aligned}
    &\mathbf{A}_{2D \rightarrow 3D}=K_{3D} V_{2D}^{\mathrm{T}}, \\
    &R_{2D \rightarrow 3D}=\operatorname{Softmax}\left(\frac{\mathbf{A}_{2D \rightarrow 3D}}{ \sqrt{F_{3D}}}\right) V_{3D}.
\end{aligned}
\label{eq7}
\end{equation}
% \vspace{-0.10cm}

%%%%%%%%% BODY TEXT - ENTER YOUR RESPONSE BELOW
\section{Evaluation}

Similar to previous 3D point cloud-based Domain Adaptation (DA) methods~\cite{jaritz2020xmuda,ganin2015unsupervised,shin2021labor}, the performance of the network is evaluated on the test set by leveraging the standard PASCAL VOC Intersection-Over-Union (IoU). The mean IoU (mIoU) is the mean of all IoU values over all classes. In detail, the mIoU can be calculated as follows:

\begin{equation}
    \mathrm{mIoU}=\frac{1}{C} \sum_{i=0}^C \frac{T P(i)}{T P(i)+F P(i)+F N(i)},
\end{equation}

\noindent where $C$ is the overall number of categories, $T P(i), F P(i)$, and $F N(i)$ are values of true positive, false positive, and false negative towards the $i$-th category, respectively.

\section{Module-level Ablation Studies.}

\begin{figure}[htbp]
  \centering
  \includegraphics[width=0.97\linewidth]{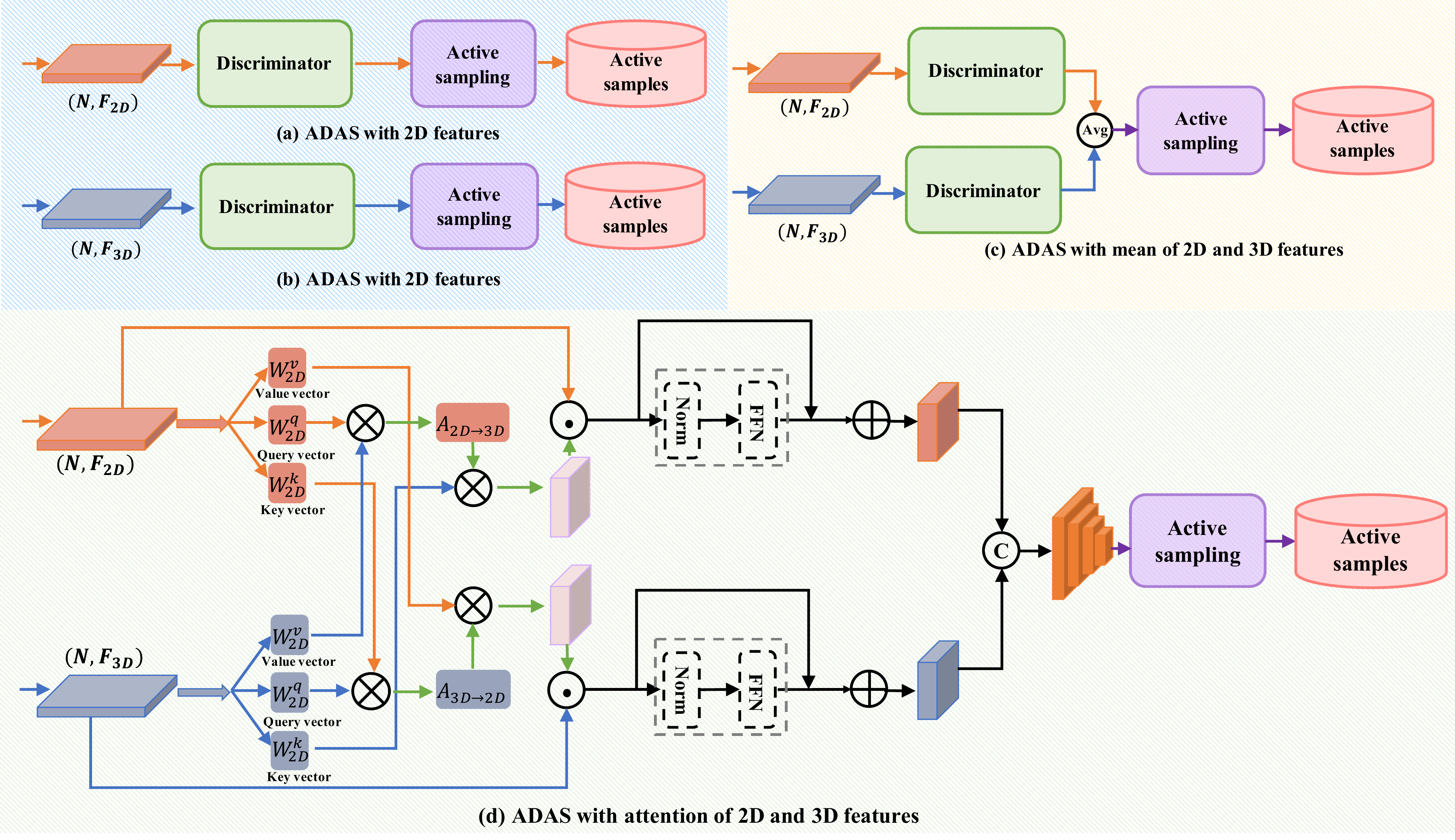}
  \vspace{0.05cm}
  \caption{The illustration of various sampling strategies: (a) Baseline 1 means that the key samples are selected only using 2D scores; (b) Baseline 2 means that the key samples are selected only using 3D scores; (c) Baseline 3 means that we sample the keyframes by means of an average across 2D and 3D scores; (d) Ours denotes the sampling scores that are calculated by the cross-attention along the 2D and 3D features.}
\label{ablation-module1}
\vspace{0.15cm}
\end{figure}

Fig.~\ref{ablation-module} and Fig.~\ref{ablation-module1} vividly depicts the comparison of the four baselines. \textbf{Baseline 1} only leverages 2D features extracted by the 2D backbone. The 2D scores obtained by the domain discriminator $\mathcal{D}$ are used to sample the source and target data. \textbf{Baseline 2} utilizes the 3D features for sampling, where the sampling process will depend on the 3D scores from the domain discriminator $\mathcal{D}$. \textbf{Baseline 3} naively averages the 2D and 3D scores for the sampling strategy. For the first two baselines, since the scores are calculated from a single modality, the sampled data might be distributed in a single modality. For instance, the data sampled by the 2D scores will concentrate more on the image information and vice versa. Compared with the above-mentioned baselines, our proposed cross-modal attention-based source-and-target sampling strategy performs an image-to-point and point-to-image feature-level information interaction by a symmetrical cross-branch attention structure. And the cross-modal features are leveraged to train a domain discriminator $\mathcal{D}$. In this way, the sampled data could take both the 2D and 3D features into consideration.

\section{More Results on 3D ADA Task}

\begin{figure}[t]
  \centering
  \includegraphics[width=0.8\linewidth]{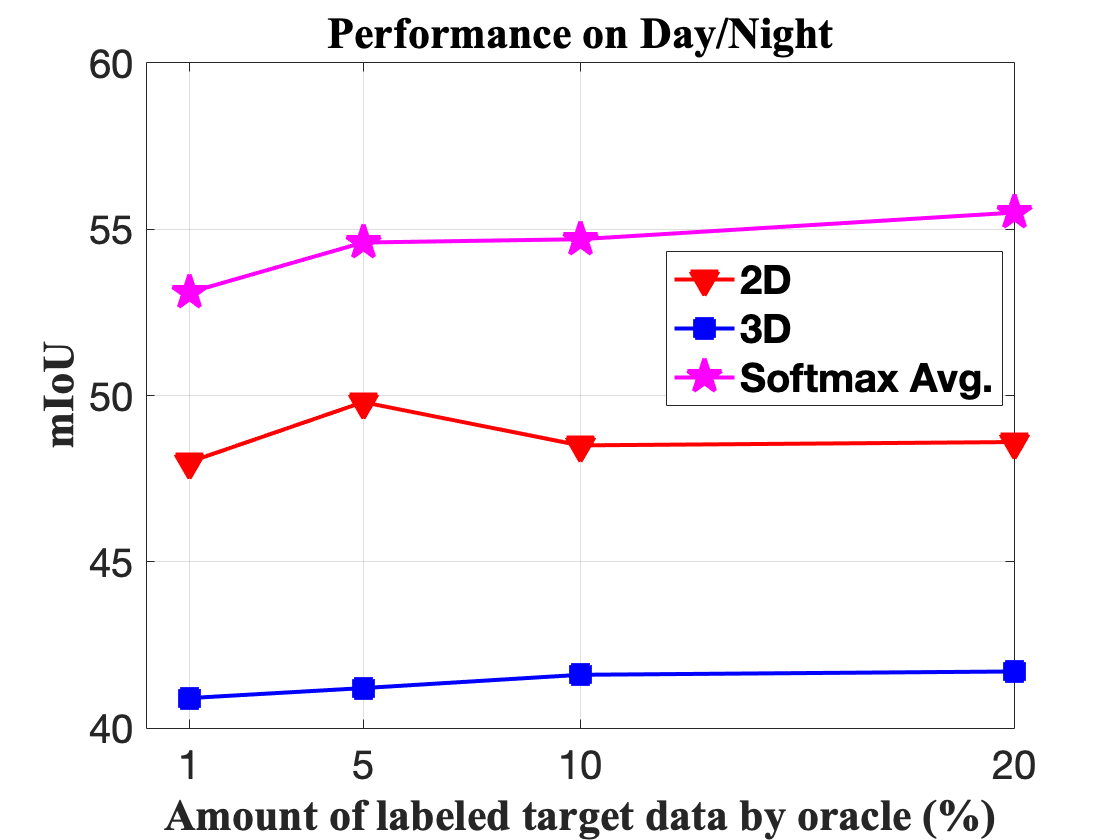}
  \vspace{0.05cm}
  \caption{The influence of different target-domain annotation budget $\mathcal{B}$ ($20\%$, $10\%$, $5\%$, and $1\%$) on the ADA task. Here, we take the Day-to-Night scenario as an example.}
\label{budget1}
\end{figure}

\begin{figure}[t]
  \centering
  \includegraphics[width=0.8\linewidth]{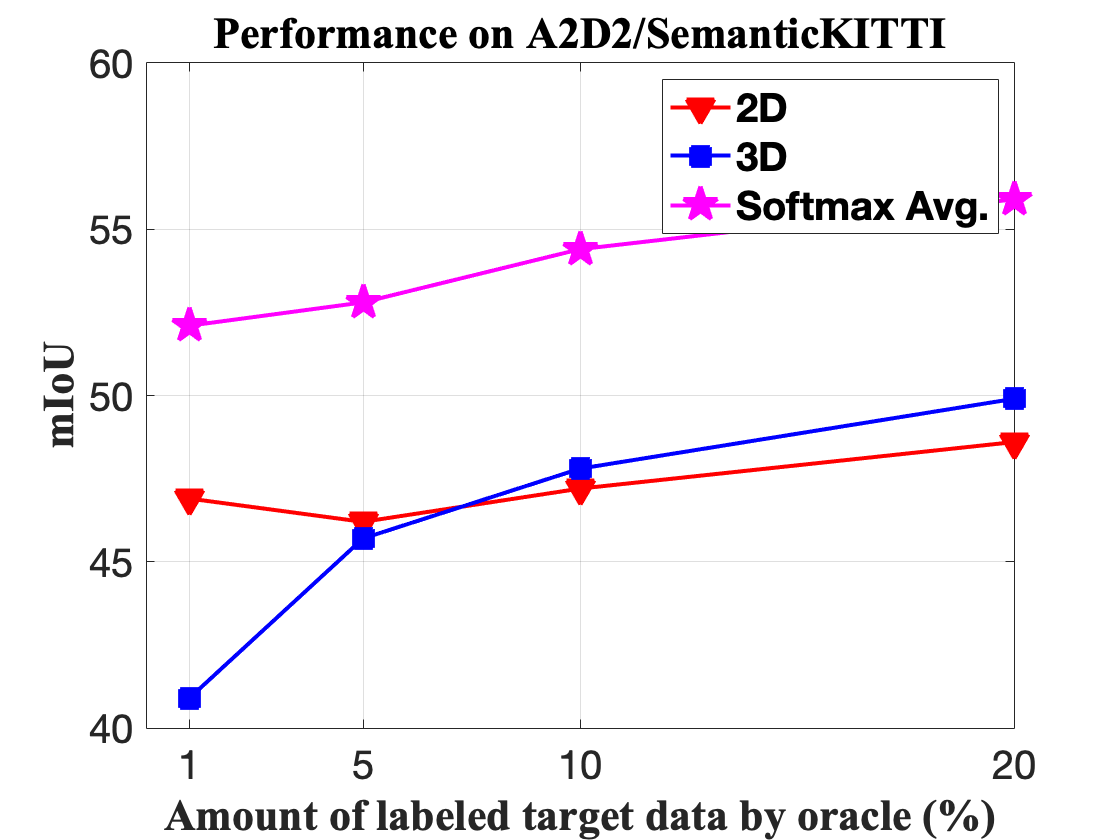}
  \vspace{0.05cm}
  \caption{The influence of different target-domain annotation budget $\mathcal{B}$ ($20\%$, $10\%$, $5\%$, and $1\%$) on the ADA task. Here, we take the A2D2-to-SemanticKITTI scenario as an example.}
\label{budget2}
\end{figure}

In addition to the performance on the ADA task, we further carry out experiments on changing the budget of the target-domain annotation under the Day/Night and A2D2/SemanticKITTI scenarios. As shown in Figs.~\ref{budget1} and~\ref{budget2}, consistent with the USA/Singapore setting in the main text, the UniDA3D also further gradually boosts the target-domain segmentation accuracy. In detail, the model performance in the target domain is improved with the increase of the annotation budget. We also observe that the 5$\%$ annotation budget can be regarded as a good trade-off between the annotation cost and the model adaptability.

\section{The Visualization of the Selected Samples}

\begin{figure*}[t]
\vspace{-0.25cm}
  \centering
  \includegraphics[width=0.82\linewidth]{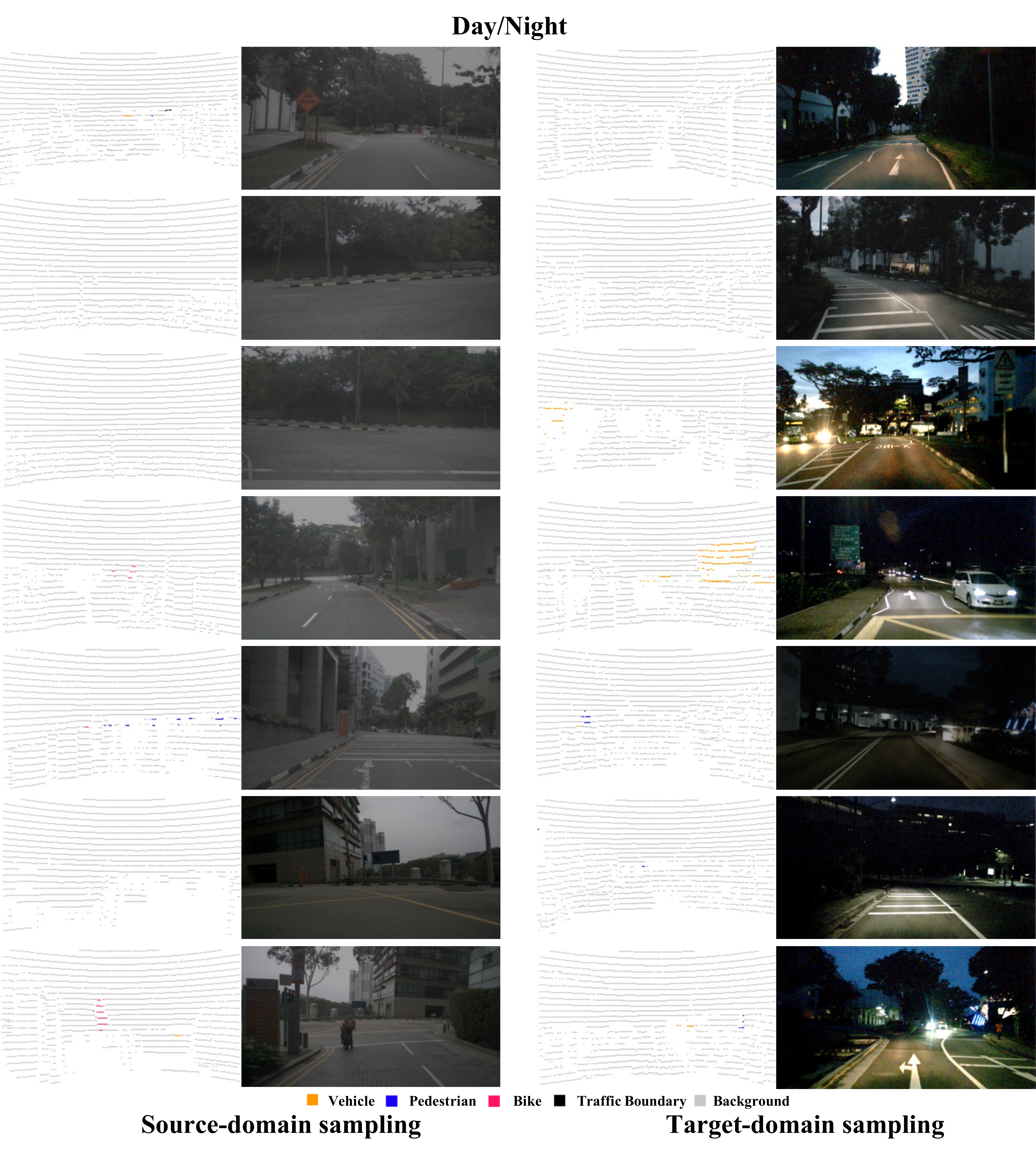}
  \caption{The samples selected by source-domain sampling and target-domain sampling under the Day/Night scenario, where the model tends to select the source-domain data at dark, since the target-domain scenes mainly cover the night scene.}
% \vspace{-0.25cm}
\label{day2night}
\end{figure*}

\begin{figure*}[t]
\vspace{-0.25cm}
  \centering
  \includegraphics[width=0.82\linewidth]{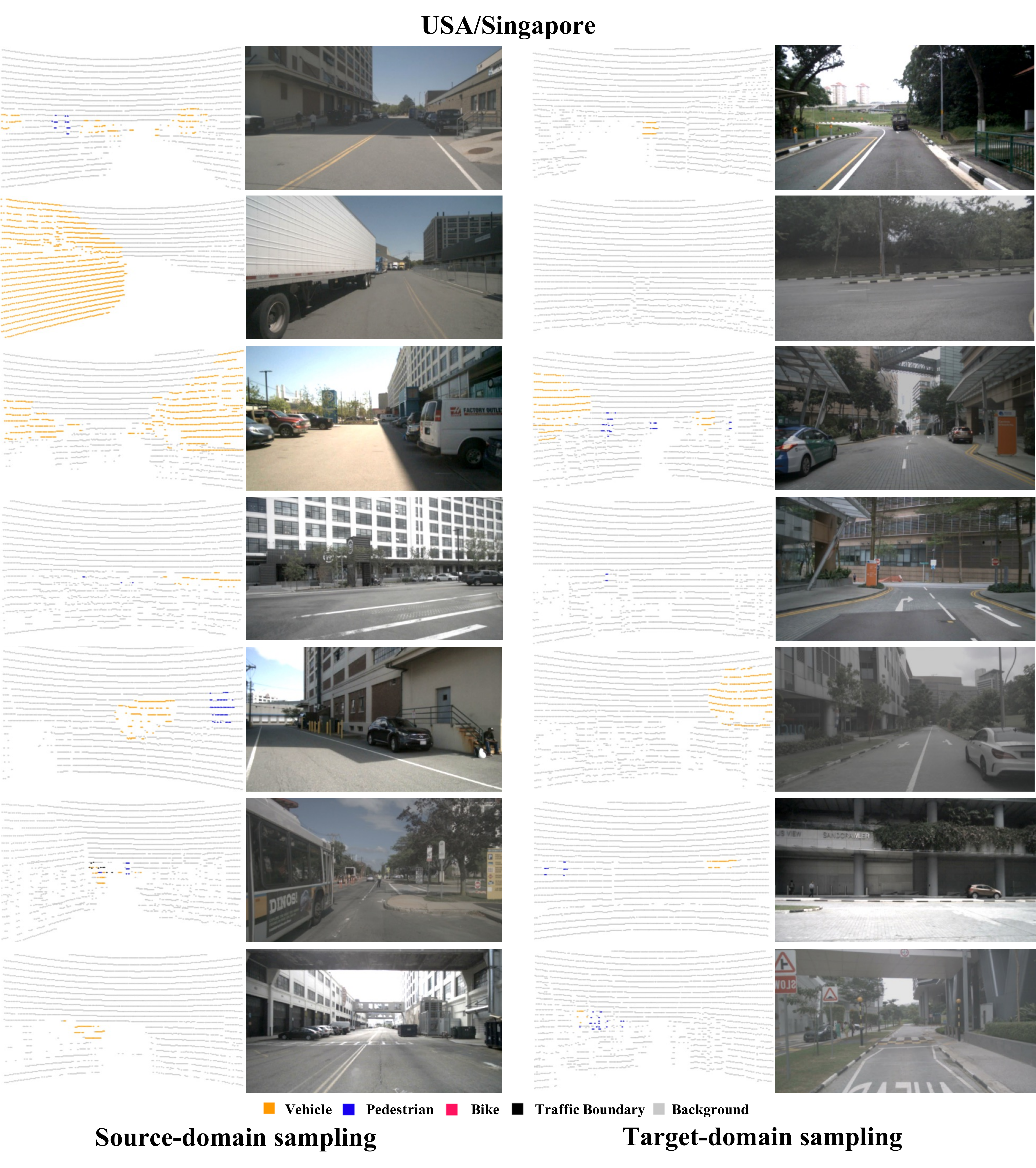}
  \caption{The samples selected by source-domain sampling and target-domain sampling under the USA/Singapore scenario, where the model tends to select the source-domain data collected from the narrow road scene.}
% \vspace{-0.25cm}
\label{usa2singapore}
\end{figure*}

\begin{figure*}[t]
\vspace{-0.25cm}
  \centering
  \includegraphics[width=0.82\linewidth]{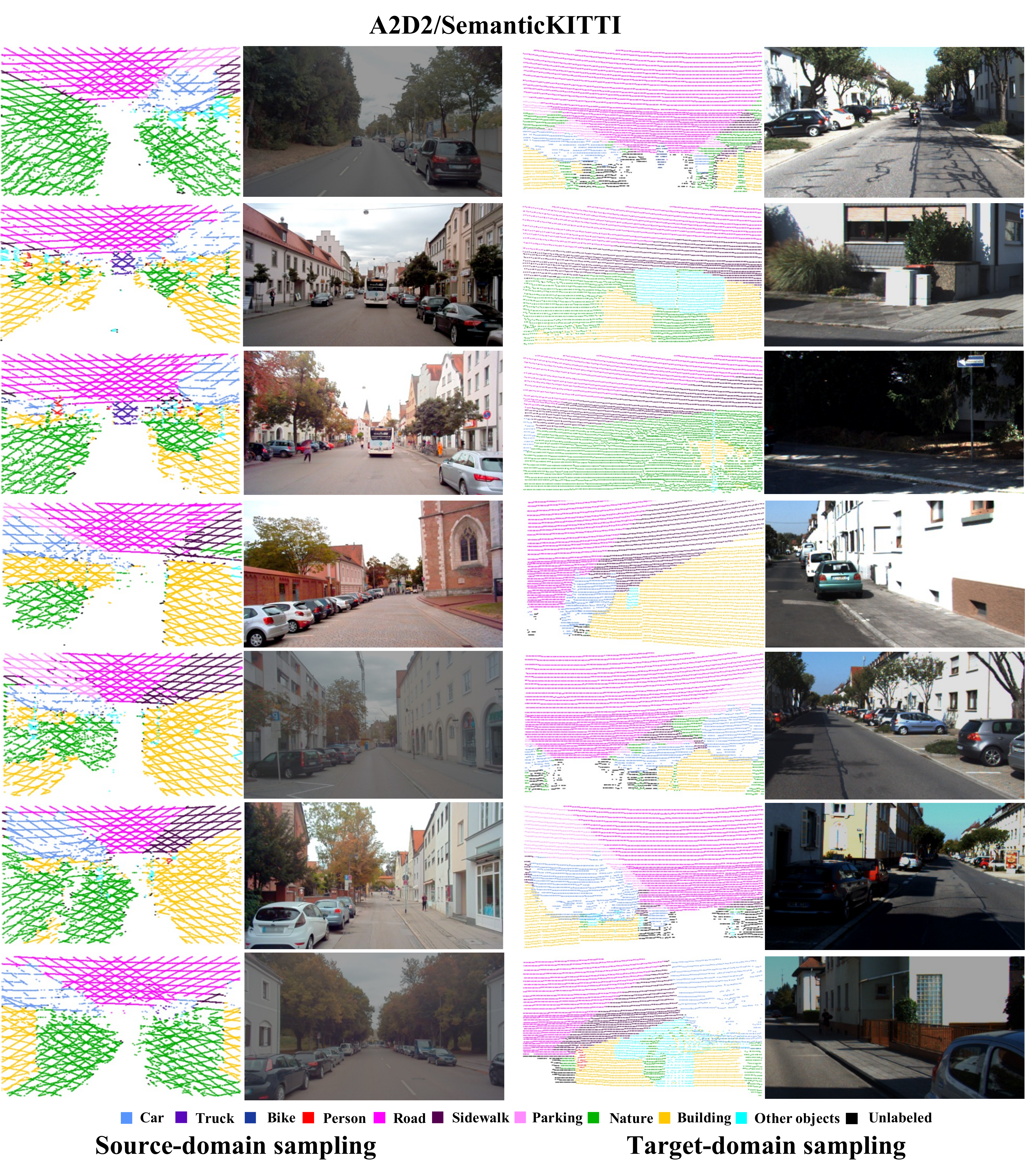}
  \caption{The samples selected by source-domain sampling and target-domain sampling under the A2D2/SemanticKITTI scenario, where for source domain, various frames that the cars parked along the road are selected by our method.}
% \vspace{-0.25cm}
\label{audi2kitti}
\end{figure*}

In order to comprehensively verify the effectiveness of the cross-modal attention-based source-and-target sampling strategy, the selected samples for both source and target domains are shown in Figs.~\ref{day2night}, ~\ref{usa2singapore}, and~\ref{audi2kitti}.

\noindent\textbf{Selected Samples from Day/Night Scenario.} As can be seen in Fig.~\ref{day2night}, the sampled source data tend to be dark, which is closer to the Night setting in the target domain. And the sampled target data is various, leading to a maximally-informative subset.

\noindent\textbf{Selected Samples from USA/Singapore Scenario.} As shown in Fig~\ref{usa2singapore}, the roads in the target domain are relatively narrow, so the source domain data selected by the source-domain sampling strategy are more inclined to the narrow roads in the USA. On the other hand, the target-domain sampling strategy also can select the representative sample.

\noindent\textbf{Selected Samples from A2D2/SemanticKITTI Scenario.} As seen in Fig~\ref{audi2kitti}, the SemanticKITTI contains various frames that the cars parked along the road. Therefore, our source-domain sampling strategy tends to choose similar frames to the target domain. The tuition is that the selected source data also contains many cars along the streets.

\section{More Results on the nuScenes Lidarseg: USA/Singapore.}

We also conduct extensive experiments on the nuScenes LiDARSeg~\cite{caesar2020nuscenes} to show the superiority of our UniDA3D. Similar to our DA experiments on the other scenarios, we perform our UniDA3D on the USA/Singapore adaptation scenario. The experimental results are shown in Table~\ref{tab:seg}. It can be observed that the baseline model trained on the source domain itself obtains $58.8\%$, $63.2\%$, and $68.5\%$ on the 2D, 3D, and softmax average performance, respectively. After the baseline is adapted to the data selected by the source-domain sampling, the segmentation results further increase to $66.6\%$, $61.0\%$, and $71.3\%$, respectively. Besides, the baseline model can be further enhanced by adapting the model to the selected target data, leading to $67.8\%$, $61.5\%$, and $72.8\%$ segmentation results. As a result, the baseline model coupled with our UniDA3D is superior to the xMUDA by $1.6\%$ and $1.7\%$ in terms of the 2D and softmax average performance, demonstrating the feasibility and strength of our UniDA3D.

\setcounter{table}{4}
\begin{table}[t]\small
\centering
\tabcolsep=0.08cm
{%
\begin{tabular}{c|ccc}
\toprule[1.5pt]
                                                                                                   nuSc-LiDARSeg: USA/Singapore  & 2D   & 3D   & Softmax avg. \\ \midrule[1pt]
Source only                                                                                          & 58.8 & 63.2 & 68.5  \\ \midrule[1pt]
xMUDA                                                                                                & 63.1 & 64.2 & 67.8  \\
xMUDA   + PL                                                                                         & 66.2 & 65.1 & 70.1  \\ \midrule[1pt]
Baseline + Source-domain sampling                                                                    & 66.6    & 61.0    & 71.3     \\
\begin{tabular}[c]{@{}c@{}}Baseline + Source-domain sampling \\ + Target-domain samping\end{tabular} & 67.8    & 61.5    & 72.8     \\ \bottomrule[1.5pt]
\end{tabular}%
}
\vspace{0.07cm}
\caption{Segmentation results (mIoU) for the UDA setting and ADA setting (5\% target-domain annotation budget) under the nuScenes-LiDARSeg USA/Singapore setting.}
\label{tab:seg}
\end{table}

% \usepackage{graphicx}
\begin{table*}[htbp]\small
\centering
{%
\begin{tabular}{cc|cc}
\toprule[1.5pt]
A2D2 class           & Mapped class  & SemanticKITTI class  & Mapped class  \\ \midrule[1pt]
Car 1                & car           & unlabeled            & ignore        \\
Car 2                & car           & outlier              & ignore        \\
Car 3                & car           & car                  & car           \\
Car 4                & car           & bicycle              & bike          \\
Bicycle 1            & bike          & bus                  & ignore        \\
Bicycle 2            & bike          & motorcycle           & bike          \\
Bicycle 3            & bike          & on-rails             & ignore        \\
Bicycle 4            & bike          & truck                & truck         \\
Pedestrian 1         & person        & other-vehicle        & ignore        \\
Pedestrian 2         & person        & person               & person        \\
Pedestrian 3         & person        & bicyclist            & bike          \\
Truck 1              & truck         & motorcyclist         & bike          \\
Truck 2              & truck         & road                 & road          \\
Truck 3              & truck         & parking              & parking       \\
Small vehicles 1     & bike          & sidewalk             & sidewalk      \\
Small vehicles 2     & bike          & other-ground         & ignore        \\
Small vehicles 3     & bike          & building             & building      \\
Traffic signal 1     & other-objects & fence                & other-objects \\
Traffic signal 2     & other-objects & other-structure      & ignore        \\
Traffic signal 3     & other-objects & lane-marking         & road          \\
Traffic sign 1       & other-objects & vegetation           & nature        \\
Traffic sign 2       & other-objects & trunk                & nature        \\
Traffic sign 3       & other-objects & terrain              & nature        \\
Utility vehicle 1    & ignore        & pole                 & other-objects \\
Utility vehicle 2    & ignore        & traffic-sign         & other-objects \\
Sidebars             & other-objects & other-object         & other-objects \\
Speed bumper         & other-objects & moving-car           & car           \\
Curbstone            & sidewalk      & moving-bicyclist     & bike          \\
Solid line           & road          & moving-person        & person        \\
Irrelevant signs     & other-objects & moving-motorcyclist  & bike          \\
Road blocks          & other-objects & moving-on-rails      & ignore        \\
Tractor              & ignore        & moving-bus           & ignore        \\
Non-drivable street  & ignore        & moving-truck         & truck         \\
Zebra crossing       & road          & moving-other-vehicle & ignore        \\
Obstacles/trash      & other-objects &                      &               \\
Poles                & other-objects &                      &               \\
RD restricted area   & road          &                      &               \\
Animals              & other-objects &                      &               \\
Grid structure       & other-objects &                      &               \\
Signal corpus        & other-objects &                      &               \\
Drivable cobbleston  & road          &                      &               \\
Electronic traffic   & other-objects &                      &               \\
Slow drive area      & road          &                      &               \\
Nature object        & nature        &                      &               \\
Parking area         & parking       &                      &               \\
Sidewalk             & sidewalk      &                      &               \\
Ego car              & car           &                      &               \\
Painted driv. instr. & road          &                      &               \\
Traffic guide obj.   & other-objects &                      &               \\
Dashed line          & road          &                      &               \\
RD normal street     & road          &                      &               \\
Sky                  & ignore        &                      &               \\
Buildings            & building      &                      &               \\
Blurred area         & ignore        &                      &               \\
Rain dirt            & ignore        &                      &               \\ \bottomrule[1.5pt]
\end{tabular}%
}
\caption{Class mapping for A2D2/SemanticKITTI scenario.}
\end{table*}

\clearpage
\newpage

%%%%%%%%% REFERENCES

% {\small
% \clearpage
% \newpage
% \vfill
% \bibliographystyle{ieee_fullname}
% \bibliography{egbib}
% }

% \input{appendix}

% add a file named appendix.tex

\end{document}